\pdfoutput=1
\documentclass{article}
\usepackage[a4paper, portrait, margin=1.1811in]{geometry}
\usepackage[english]{babel}
\usepackage[utf8]{inputenc}
\usepackage[T1]{fontenc}
\usepackage{helvet}
\usepackage{etoolbox}
\usepackage{graphicx}
\usepackage{titlesec}
\usepackage{caption}
\usepackage{booktabs}
\usepackage{xcolor}

\usepackage{hyperref}
\hypersetup{colorlinks,citecolor=cyan}

\usepackage{caption}
\usepackage{apacite}
\usepackage{afterpage}
\usepackage{algorithm}
\usepackage[noend]{algpseudocode}
\usepackage[utf8]{inputenc}
\usepackage{amsmath}
\usepackage{bm}
\usepackage{bbm}

\usepackage{wrapfig}
\DeclareMathOperator*{\argmax}{arg\,max}

\usepackage{scrextend}
\usepackage{fancyhdr}
\usepackage{graphicx}
\newcounter{lemma}
\newtheorem{lemma}{Lemma}
\newcounter{theorem}
\newtheorem{theorem}{Theorem}

\fancypagestyle{plain}{
	\fancyhf{}

}

\makeatletter
\patchcmd{\@maketitle}{\LARGE \@title}{\fontsize{16}{19.2}\selectfont\@title}{}{}
\makeatother

\usepackage{authblk}

\newsavebox\affbox
\author[1$\ast$]{\textbf{Yanli Zhou}}
\author[2]{\textbf{Reuben Feinman}}
\author[1,3]{\textbf{Brenden M. Lake}}

\affil[1]{ Center for Data Science, New York University
}
\affil[2]{ Center for Neural Science, New York University
}
\affil[3]{ Department of Psychology, New York University
}

\titlespacing\section{0pt}{12pt plus 4pt minus 2pt}{0pt plus 2pt minus 2pt}
\titlespacing\subsection{12pt}{12pt plus 4pt minus 2pt}{0pt plus 2pt minus 2pt}
\titlespacing\subsubsection{12pt}{12pt plus 4pt minus 2pt}{0pt plus 2pt minus 2pt}

\titleformat{\section}{\normalfont\fontsize{10}{15}\bfseries}{\thesection.}{1em}{}
\titleformat{\subsection}{\normalfont\fontsize{10}{15}\bfseries}{\thesubsection.}{1em}{}
\titleformat{\subsubsection}{\normalfont\fontsize{10}{15}\bfseries}{\thesubsubsection.}{1em}{}

\titleformat{\author}{\normalfont\fontsize{10}{15}\bfseries}{\thesection}{1em}{}

\title{\textbf{\huge Compositional diversity in visual concept learning}\\
}
\date{}

\usepackage{wrapfig}

\usepackage{color}
\newcommand\todo[1]{%
  \bgroup
  \hskip0pt\color{red!80!black}%
  TODO: #1%
  \egroup
}

\begin{document}

\pagestyle{headings}	
\newpage
\setcounter{page}{1}
\renewcommand{\thepage}{\arabic{page}}

\captionsetup[figure]{labelfont={bf},labelformat={default},labelsep=period,name={Figure }}	\captionsetup[table]{labelfont={bf},labelformat={default},labelsep=period,name={Table }}
\setlength{\parskip}{0.5em}
	
\maketitle
	
\noindent\rule{15cm}{0.5pt}
	\begin{abstract}
		Humans leverage compositionality to efficiently learn new concepts, understanding how familiar parts can combine together to form novel objects. In contrast, popular computer vision models struggle to make the same types of inferences, requiring more data and generalizing less flexibly than people do. Here, we study these distinctively human abilities across a range of different types of visual composition, examining how people classify and generate ``alien figures'' with rich relational structure. We also develop a Bayesian program induction model which searches for the best programs for generating the candidate visual figures, utilizing a large program space containing different compositional mechanisms and abstractions. In few shot classification tasks, we find that people and the program induction model can make a range of meaningful compositional generalizations, with the model providing a strong account of the experimental data as well as interpretable parameters that reveal human assumptions about the factors invariant to category membership (here, to rotation and changing part attachment). In few shot generation tasks, both people and the models are able to construct compelling novel examples, with people behaving in additional structured ways beyond the model capabilities, e.g. making choices that complete a set or reconfiguring existing parts in highly novel ways.  To capture these additional behavioral patterns, we develop an alternative model based on neuro-symbolic program induction: this model also composes new concepts from existing parts yet, distinctively, it utilizes neural network modules to successfully capture residual statistical structure.  Together, our behavioral and computational findings show how people and models can produce a rich variety of compositional behavior when classifying and generating visual objects.\\ \\
		\let\thefootnote\relax\footnotetext{
			\small $^{*}$\textbf{Corresponding author.} \textit{
				\textit{E-mail address: \color{cyan}yanlizhou@nyu.edu}}\\
		}
		\textbf{\textit{Keywords}}: \textit{concept learning; Bayesian inference; few-shot learning; visual learning; compositionality; neuro-symbolic models}
	\end{abstract}
\noindent\rule{15cm}{0.4pt}

\section*{Introduction}

\begin{figure}[ht]
\centering
\includegraphics[width=0.85\linewidth]{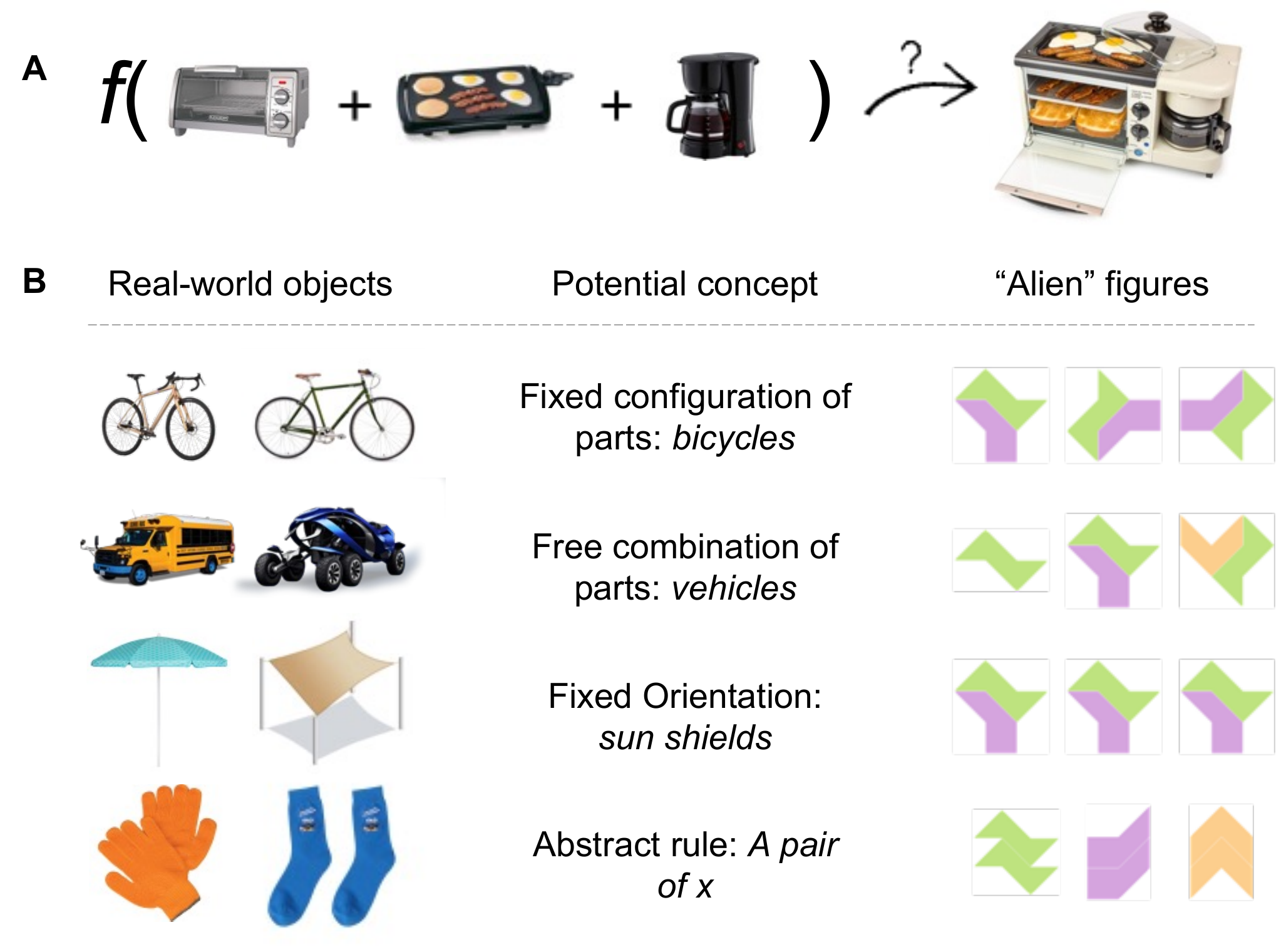}
\caption{\textbf{Visual concept learning requires combining familiar parts in a diverse range of ways.} (A) Humans can learn the concept of \emph{breakfast machine} with a single example by recognizing familiar components and reasoning about their relations. Leading computer vision models tend to struggle with this concept. (B) Real-world visual concepts are defined by different types of compositions: 1. A \emph{bicycle} is a well-defined collection of parts in a consistent configuration; 2.  \emph{vehicles} allow a set of stereotyped parts to be combined more freely; 3. To be a \emph{sun shield}, an upright orientation is required; 4. \emph{A pair of $x$} stipulates a repetition of \emph{wearable} elements. The rightmost column contains examples of experimental stimuli that are analogous to these concepts.}
\label{intro}
\end{figure}

Compositional generalization, the reuse and recombination of preexisting knowledge to handle novel cases, is a cornerstone of human intelligence. Generalization in natural language is a quintessential example: people can understand and generate infinitely many sentences from a finite number of words \cite[quoting Wilhelm von Humbolt]{chomsky_1957, chomsky_1965}. Generalization in visual cognition can be characterized similarly: people can understand a potentially infinite number of scenes through combinations of objects, or learning about new objects through combinations of parts and relations. For example, people who are familiar with \emph{coffee maker}, \emph{toaster oven} and \emph{griddle} can effortlessly grasp the concept of a \emph{breakfast machine} upon seeing it for the first time (Fig.~\ref{intro}A)\footnote{Example from Vicarious Research Blog.}. By recognizing the known components and reasoning about how these components are compositionally related, a learner can formulate hypotheses that accurately generalize to future encounters with other breakfast machines. 

% The ability to generalize compositionally supports rapid learning from a very young age. Early in development, children can make meaningful generalizations from exposure to just one or few positive examples of a new concept (\shortciteNP{smith_object_2002}; \citeNP{xu_word_2007}). 

In contrast to people, today's computer vision models, while successful in many applications, are limited in their abilities to make compositional generalization and learn from very few examples. Children can make meaningful generalizations from one or few positive examples of a new concept \shortcite{smith_object_2002,xu_word_2007}. Compared to people, neural network systems, while advancing in their few-shot learning abilities \shortcite{Hospedales2022,LakeOmniglotProgress}, typically require more data and require more task-specific training \shortcite{lake_building_2017}. Recent multi-modal models that combine images and text, such as text-to-image generative models, can make impressive compositional generalizations in some cases (``a teddybear on a skateboard in times square'') and then fail in other related cases (``a red cube on top of a blue cube'') \shortcite{Ramesh2022}. For instance, a strong image captioning system \cite{blip} describes the breakfast machine in Fig.~\ref{intro}A as \emph{``toaster oven with toast on the bottom and breakfast fried egg"}, identifying some of the key parts while misunderstanding the larger compositional whole. Understanding human compositional visual concept learning in computational terms is therefore a natural step to building machine learning models that can harness compositionality more like people do.

To study how people learn and represent visual concepts in computational terms requires us to first recognize the qualitatively different types of composition present in our visual world (see examples in Fig.~\ref{intro}B). A concept like \emph{bicycle} stipulates a fixed configuration of parts and their relations (e.g.\ bikes have handlebars, a seat, and two wheels in a consistent configuration), whereas a concept like \emph{vehicle} allows category members to have freer combinations of parts and relations (varying numbers of wheels, motors, etc. are acceptable). A concept like \emph{sun shield} requires selectivity of object orientation, in order to fulfill a given conceptual constraint. Finally, a concept like \emph{a pair of x} requires an additional degree of compositional abstraction, allowing a variety of parts to fill a role as long as they are duplicated. Forming a comprehensive understanding of the different visual composition types poses a challenging learning problem that requires manipulating parts and relations at various levels of abstraction.

In this work, we take on the challenge of studying how people learn concepts that utilize a rich diversity of part-based composition, as present in real-world visual concepts, and developing a unifying computational model that can capture these different types of compositional generalization. To achieve this, our strategy for building computational models brings together three ingredients that have been influential in previous research on few-shot concept learning. The first ingredient is Bayesian modeling, which allows for the incorporation of prior knowledge and for explicit reasoning over generative hypotheses \shortcite{Tenenbaum1999,xu_word_2007,Tenenbaum2011}. The second ingredient is a structured description language for visual concept learning \shortcite{stuhlmuller_learning_2010,lake_human-level_2015,overlan_learning_2017,lake_recursive}, relating to the types of grammars and formal languages used for modeling compositionality in natural languages \shortcite{ChierchiaGennaroandMcConnell-Ginet1990}, or to computer programming and formal logic that are perfectly systematic in how expressions combine. The third ingredient is the utilization of powerful neural network modeling components, as instantiated through hybrid neuro-symbolic modeling \shortcite{Hewitt2020, Ellis2021, Kulkarni2015, Feinman2021}. This modeling approach uses both neural networks and symbolic representation to amortize, accelerate and improve upon more purely symbolic or neural models.

Previous empirical studies and modeling efforts, while providing essential guidance through the above ingredients, have largely been restricted to special cases of visual compositionality, in contrast to the broader scope we aim for here. For instance, \citeA{xu_word_2007}'s work on Bayesian word learning helps to explain how children can make meaningful inferences from just a few examples, but their model operates over a hypothesis space that treats objects as unified wholes rather than compositions of parts. The class of handwritten characters considered in \citeA{lake_human-level_2015} and \citeA{Feinman2021} is inherently compositional, but individual characters are highly constrained in how parts and configuration are allowed to vary (as in the 1\textsuperscript{st} row of Fig.~\ref{intro}B). The sequential patterns studied in \citeA{overlan_learning_2017} and \citeA{LakeLinzenBaroni2019} and recursive structures in \citeA{stuhlmuller_learning_2010} and \citeA{lake_recursive} are special case studies more akin to the 4\textsuperscript{th} row of Fig.~\ref{intro}B. The free combinations of parts arranged in grid-like scenes in \citeA{orban2008} are most analogous to the 2\textsuperscript{nd} row of Fig.~\ref{intro}B. Each of these case studies also considered only relatively simple types of spatial relations.
% Prior studies on neuro-symbolic models have either focused on a single class of generalization \cite{Feinman2021} or have used neural networks solely for inference in a fully-symbolic generative model \shortcite{Ellis2021, Kulkarni2015}.
Our goals differ in that we would like to account for multiple types of visual compositional generalization within a single experimental paradigm and computational framework.
% , and in the spirit of neuro-symbolic modeling, we also demonstrate how neural networks can be used in forward generative models to improve the fit to human behavioral data.

To achieve a sufficiently powerful experimental paradigm, we develop a domain of visual concepts that is richly hierarchical, compositional, and relational, which we call the ``alien figures" domain. The class of alien figures is capable of representing various composition types ranging from fixed spatial relations like \emph{bicycles} to abstract patterns like \emph{pairs of x} (examples in Fig.~\ref{intro}B, right column). Using alien figures as our stimuli, we first conduct behavioral experiments on few-shot concept learning, asking participants to classify novel visual figures after observing just a few positive examples of a novel visual class. For human concept learning, the ability to classify new examples comes with other abilities too, including the second ability we focus on in this article: generating new examples. Generative paradigms are especially rich in terms of eliciting complex human behavior \shortcite{Ward1994,Jern2013,lake_human-level_2015}, and thus in a second experiment, we ask participants to generate novel examples based on a few examples of a novel concept. We test a wide variety of composition types in both categorization and generation experiments and document a suite of interesting behavioral patterns. For instance, we observe a strong assumption for invariance of object orientation and part attachments that persists throughout different tasks, and a distinct inductive bias we termed as ``complete-the-pattern", which is characterized by an overwhelming preference for selecting a specific orientation or part for generation if a pattern can be completed. Detailed descriptions of all inductive biases are provided in later sections.

To develop a unifying computational model that can account for the inferences people make when presented with different composition types, we utilize the Bayesian program induction framework for searching for the best casual generative process for explaining a given set of visual exemplars \shortcite{stuhlmuller_learning_2010,lake_human-level_2015,overlan_learning_2017,lake_recursive}. Specifically, a hypothesis regarding the meaning of a visual concept is operationalized as a probabilistic program that, when run, produces a set of category exemplars as output. To construct the (potentially infinite) set of possible visual concepts, we design a probabilistic grammar that produces an unbounded set of visual concepts from a small set of primitive operations \cite{goodman_rational_2008,piantadosi_2011,piantadosi_four_2016}. This grammar defines a domain specific language for expressing compositional visual concepts, and the probabilistic nature of the grammar allows for expressing prior expectations about which types of concepts are more likely. Crucially, the unique geometric properties of our shape primitives create spatial arrangements between object parts beyond simplistic relations such as \textit{above}, \textit{below}, \textit{left-of}, and \textit{right-of}. The domain specific language also supports variable abstraction and manipulation \cite{Marcus2003}, as utilized for representing concepts defined through abstract rules such as repeated part structure (Fig.~\ref{intro}B last row). Together the space of programs encompasses a wide range of compositions and abstractions we are interested in studying.

Under the Bayesian program induction framework, learning a new visual concept amounts to a search for the best programs for explaining the examples (here, the alien figures) under a Bayesian score. We find that our Bayesian program induction model provides a strong account of experimental data in both the categorization and generation tasks, outperforming alternative models that lack key capacities to represent relations and compositionality. Furthermore, the fitted model parameters are psychologically meaningful, each representing the strength for generalization preferences such as orientation invariance, providing insight into a number of people's inductive biases for these few-shot learning tasks.

Finally, despite its explanatory power, we find that the Bayesian program induction model is not a perfect account of the human behavior. Upon close inspection, human behavior can deviate from the model in ways defy simple symbolic description or specification. Motivated to account for these additional behaviors, we also utilize a more data-driven, neuro-symbolic approach (ingredient 3) to model building structured probabilistic models (ingredients 1 and 2), following recent work on generative neuro-symbolic (GNS) modeling \cite{Feinman2021}. Like the Bayesian program inductive model, this approach posits human concepts as probabilistic programs for generating new examples; however, GNS uses powerful neural network estimators, in conjunction with a tailored meta-learning scheme, to capture the complex statistical structure underlying human generalization that might evade a fully-symbolic probabilistic model. As a result, GNS can provide a more comprehensive behavioral account while offering much of the same structure and interpretability.

\section{Experiment 1: Few-shot categorization of compositional visual concepts}

\begin{figure}[ht]
\centering
\includegraphics[width=0.85\linewidth]{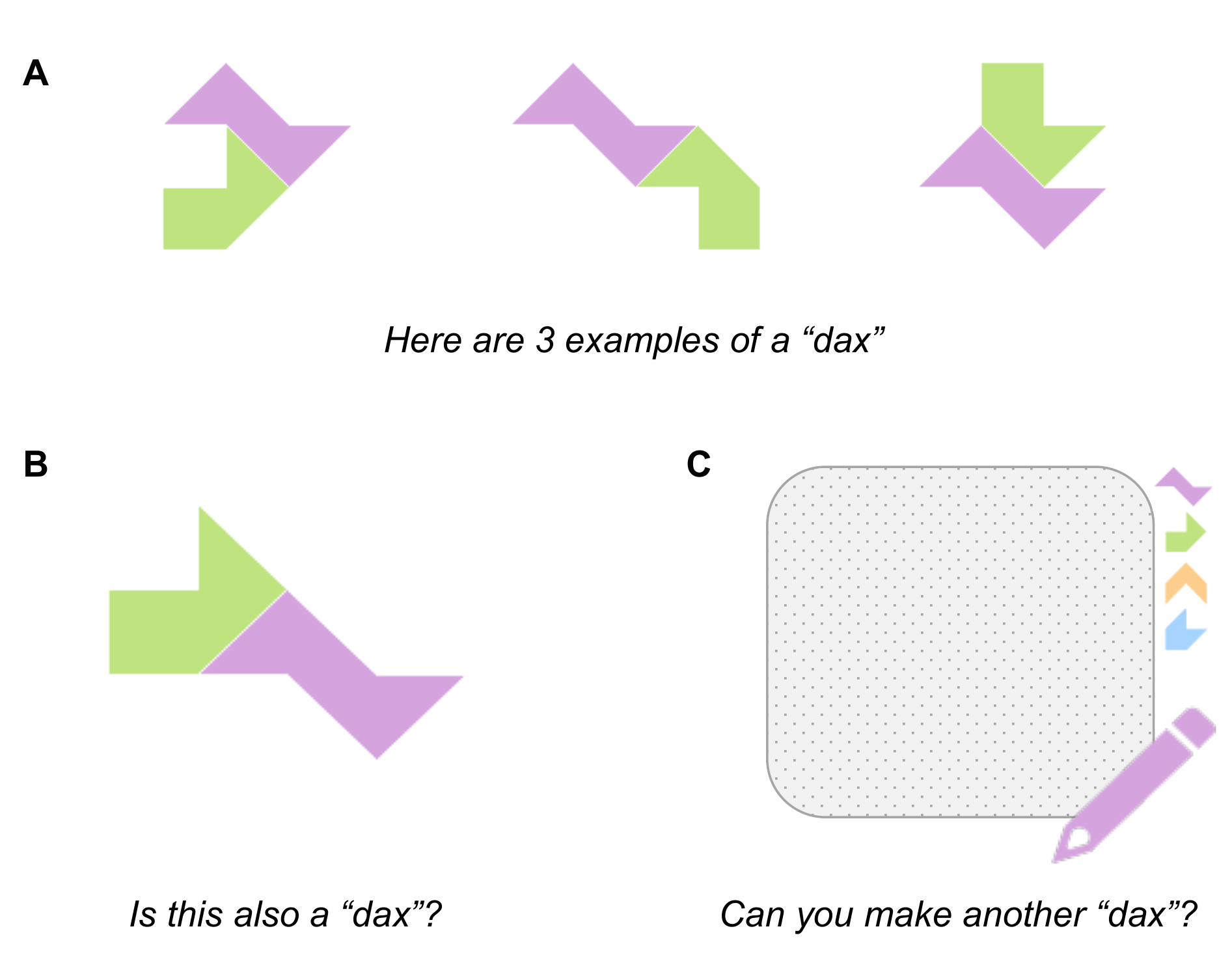}

\caption{{\bf Example trial for learning a novel visual concept.}
(A) In all experiments, the participant is first familiarized a set of exemplars from a novel visual concept. (B) In the categorization task, the participant is then presented with a series of novel figures and judges whether each belongs to the same concept as the provided exemplars. (C) In the generation task, the participant is instead given a digital interface that allows them to generate a novel example of the candidate concept.}
\label{procedure}
\end{figure}

\begin{figure*}[ht]
  \includegraphics[width=\textwidth]{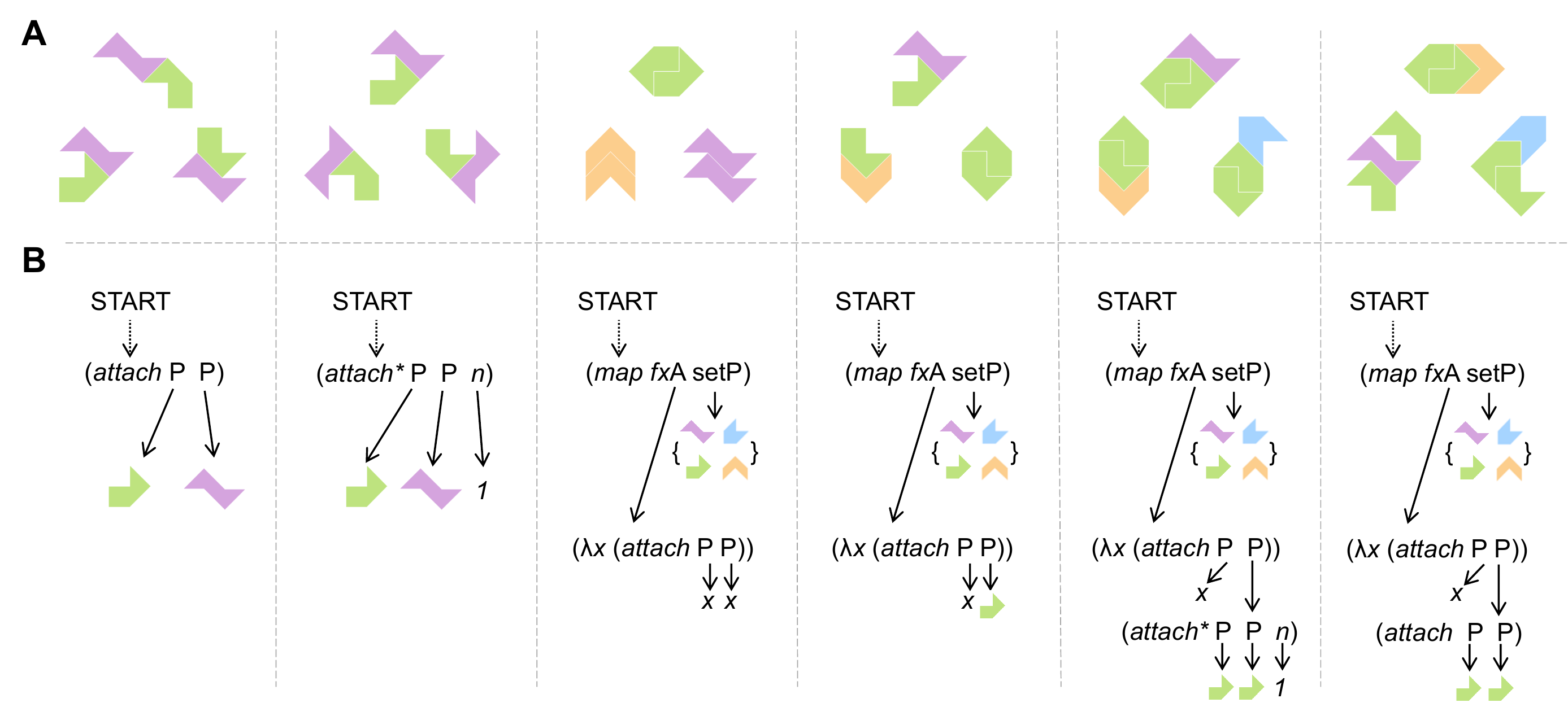}
  \caption{\textbf{Examples of trial types tested in the experiments.} (A) Sets of example alien figures given to participants to study on a given trial. (B) Simplified parse tree of the most likely concept inferred by the Bayesian program induction model for each trial. The grammar over programs specifies primitive shapes and operations including the \emph{attach} function, which returns the set of all possible configurations of two parts, and the \emph{attach*} function which returns a set containing the single specified configuration ($1^{st}$ allowable attachment). We can see that the spatial arrangement among components cannot be described with simple relations such as \emph{left-of}, \emph{right-of}, \emph{above} or \emph{under}. In all of these examples, we omit a rotation function that was applied to all produced configurations as none requires a specific rotation.}
\label{concept}
\end{figure*}

\subsection{Behavioral experiments}
In a series of few-shot categorization experiments, we aim to evaluate the flexibility of human compositional learning across a range of concept types using a novel class of visual stimuli. Our task design, which is described below, builds upon the design used in the seminal work by \citeauthor{xu_word_2007} \citeyear{xu_word_2007}.

\subsubsection{Stimuli} 
 The stimuli were described to participants as ``alien figures,'' which were programmatically generated by composing one to three shape primitives (see examples in Fig.~\ref{procedure}).  A composition of two shape primitives is considered valid when they are non-overlapping and connected via two sides of identical length. The primitives themselves were constructed through an additional degree of compositionality, as they were composed of four isosceles right triangles, which gives rise to non-canonical forms that are not easily associated with common shape categories to reduce potential priors. Note that participants in the experiment saw the primitives as black-and-white outlines rather than shapes filled with color. We left these primitives uncolored to motivate closer observations of the stimulus shapes. As a visual aid in the experiment, rolling one's mouse over a primitive led all identical primitives in the display to become highlighted. (In all figures in this article, the shape primitive types are color-coded as a proxy for this roll-over functionality that participants utilized.)
 
 Trials are designed to span a wide range of compositionality types. To form the set of training examples for each trial, we varied (1) which primitives can appear, (2) how many primitives appear in each exemplar (3) how the parts are composed and (4) if the configuration has a fixed orientation (see Fig.~\ref{concept}A for examples of different trial types).

\subsubsection{The classification task} 
Participants took part in an online ``alien figure categorization game" in which they were the assistant to a professor who collected samples of alien figures on a newly discovered planet. Their job in the game was to help the professor categorize a series of unnamed alien figures based on a small set of named examples. 

During each trial, participants were first familiarized with 4 different shape primitives. They were also informed that all relevant figures within the trial were built from these 4 primitives and no other primitives were possible. Next, participants were given a small set of example figures that shared a common name (see Fig.~\ref{procedure}A for an example trial). To minimize the effect of memory demands on learning, a display of the examples and primitives remained on screen throughout the trial. After an untimed observation period, participants entered a test stage in which they categorized a series of 9-13 unnamed alien figures (see Fig.~\ref{procedure}B for an example test item). Specifically, participants chose `yes' or `no' for each test image to indicate whether it belongs to the same named category as the example images. We constructed each test set to cover a wide range of both possible and impossible extensions of potential concepts related to the training examples.

We conducted two separate experiments with identical task procedures. The two experiments differed only in terms of the training and test sets in each trial. In Experiment 1a, for every participant we tested 11 trials with each trial containing 1 to 3 training examples, followed by judgments on a set of 9 to 13 test examples. Experiment 1b consisted of 10 trials and considered concept types that were more complex compared to those used in Experiment 1a. To study the effect of the exemplar set size on learning, participants in Experiment 1b were randomly separated into two conditions, based on whether they saw 3 or 6 exemplars of each concept. Trial orders were randomized for each participant. For each trial, we pre-generated 5 random sets of candidate primitives, and the primitive assignment was randomly sampled from the 5 for each participant. In all subsequent analyses, we combined data collected from both experiments into what we refer to as the categorization dataset, as both experiments shared an identical setup. 

\subsubsection{Participants} 

For both experiments, participants (total $N=100$) were recruited via Amazon's Mechanical Turk. In Experiment 1a, 40 participants took part and in Experiment 1b, 30 participants took part in each condition. We implemented an attention check on every trial by asking participants to indicate whether one of the exemplars belongs to the concept. Responses from participants that failed one or more attention checks during either experiment were excluded. In the end, generalization judgments from 32, 25, and 20 participants were used in our reported analyses of Experiment 1a, the 3-exemplar condition of Experiment 1b, and the 6-exemplar condition of Experiment 1b, respectively. Participants took 47.2 minutes on average to finish the task, and were paid \$5.00 at the completion of the experiment.

\subsection{Computational models}

We explore several types of computational models, with the aim of characterizing human categorization judgments in the alien figure task in computational terms. This section introduces the Bayesian program induction model with strong compositional abilities, as well as alternative models that we hypothesize lack key aspects of compositionality necessary for capturing human behavior. The generative neuro-symbolic model is considered in Experiment 3.

% \begin{figure}[t]
%   \includegraphics[width=\linewidth]{grammar.pdf}
%   \caption{Core grammatical rules used to generate concept programs. The hypothesis space used in the study consisted of valid compositions of these primitives. Full grammar and supplementary material will be available online: https://github.com/yanlizhou/AlienFigures.}
% \label{grammar}
% \end{figure}

\subsubsection{Bayesian program induction} \label{sec_BPI}

We develop a Bayesian program induction model that considers explicit, structural hypotheses as explanations for sets of visual exemplars. Specifically, the hypotheses are alien concepts represented as probabilistic programs, which are generative models that produce distributions of examples. Inspired by previous probabilistic language of thought models in cognitive science \shortcite{goodman_rational_2008,piantadosi_2011,piantadosi_logical_2016}, we form a compositional hypothesis space using a probabilistic grammar. The grammar defines a set of primitive visual parts and primitive functions, and together these primitives can be structurally combined to build up programs of various levels of complexity (see Fig.~\ref{concept} for examples of programs and output). The production rules of the grammar specify the infinite space of possible concepts; each sample from the probabilistic grammar corresponds to a potentially different visual concept.

The goal of the learner is to infer the most probable programs under a Bayesian score---that is, the programs most consistent with the observed set of example alien figures and the prior beliefs over programs. Specifically, given a set of examples $X=\{x_1,\dots,x_k\}$, the learner aims to find the best programs $h$ according to the posterior probability, 
\begin{equation}
P(h|X) \propto P(h)P(X|h).
\end{equation}
We define the prior probability of a concept $P(h)$ and the likelihood of a concept given observation $P(h|X)$ in the following sections. 

\subsubsection{Prior over programs} Following \citeA{goodman_rational_2008}, the prior is operationalized through a probabilistic context-free grammar (PCFG) that we denote as $G$ (see Appendix Fig.~\ref{fullgrammar} for the full set of grammatical rules). To generate a concept, our grammar $G$ begins with expanding the START symbol into downstream nodes according to applicable rewrite rules. These nodes are subsequently rewritten until no further expansions are possible. The output of each program is the set of all possible alien figures under the concept. In the example \texttt{(\emph{rotate\textsuperscript{*}} (\emph{attach\textsuperscript{*}} p\textsubscript{2} p\textsubscript{4}, 1), 180)}, the inner most expression is first evaluated and returns the $1^{st}$ allowable configuration of the specific two primitives $p_2$ and $p_4$. All possible configurations of any two parts defined in the study were fully enumerated and stored, such that each configuration ID corresponds to a specific configuration. The inner part then gets passed on to the outside expression that generates a rotated copy at $180^{\circ}$. This program has only a single element in its output set, as it corresponds to a generative process that fully specifies the types of parts, their configuration, and overall rotation. Figure orientation is based on four discrete possibilities, and two identical configurations at different rotations are considered distinct alien figures.  

The grammar also supports $\lambda$-expressions: together with mapping and set operations, the grammar can produce abstract concepts like \texttt{(\emph{map} (lambda x (\emph{attach} x x)) S)} which outputs the set of all possible configurations of two identical components sampled from the set $S$. Other function primitives in the grammar support hypotheses that do not fully specify a composition process, but rather identify one or more defining parts. For example, \texttt{(\emph{has} p)} returns the set of all possible alien figures with $p$ as a part.

Formally, each node in $G$ is either a nonterminal $A$ or a terminal, both are a return type of some primitive function defined in $G$. A nonterminal $A$ is expanded into downstream nonterminals until a terminal is reached, at which point no further expansion will take place. The grammar $G$ also defines a set of rules on how its primitives can be combined. Each rule in $G$ has an associated probability, and together these probabilities are formalized with parameters $\vec{\theta}$ which quantify the distribution of expansions for the nonterminals. Therefore, the prior distribution is defined by modelling each non-terminal $A \in G$ as a multinomial, parameterized by $\vec{\theta}_A$, the set of expansion probabilities associated with all possible options of $A \rightarrow B$, such that $\sum_{B} \theta_{A \rightarrow B} = 1$. As a result, the prior probability of a hypothesis $h$ is simply the product of all production probabilities $\theta_{A \rightarrow B}$ associated with all relevant expansions $A \rightarrow B$ in $h$:
\begin{equation} \label{prior}
\begin{split}
P(h; \vec{\theta}) = \prod_{A \rightarrow B \in h} \theta_{A \rightarrow B}.\
\end{split}
\end{equation}
This formulation operationalizes an important psychological preference for simplicity \cite{chater_simplicity:_2003} as shorter programs require fewer multiplications of expansion probabilities.

\subsubsection{Likelihood} The likelihood of $X$ assuming hypothesis $h$ is true is defined as 
\begin{equation} \label{ll}
\begin{split}
P(X|h) = \prod^k_{i=1} P(x_i|h) = \prod^{k}_{i=1} \mathbbm{1}(x_{i} \in h) \cdot \frac{1}{|h|},
\end{split}
\end{equation}
where $\mathbbm{1}(\cdot)$ is the indicator function indicating whether the $i$th exemplar $X_{i}$ is a valid token of $h$, and $|h|$ is the size of the given hypothesis $h$, represented by the number of all possible tokens under $h$. For example, the left-most concept depicted in Fig.~\ref{concept} is a program that produces the set of all possible tokens in which each token is a validly attached configuration of the two participating shape primitives. The size of this concept is then equal to the set size of all unique output tokens of the program. A likelihood function that is inversely proportional to the concept size reflects the psychologically important \emph{size principle}, which assigns more weight to more specific hypotheses \cite{Tenenbaum1999}.

\subsubsection{Categorization decisions and approximate Bayesian inference} To generate a model prediction for each test item $y$ after making a set of observations, we calculate the probability that the label $l_y \in \{0,1\}$ of $y$ is consistent with the set of observed examples $X$ as 
\begin{equation} \label{response1}
\begin{split}
P(l_y = 1 | X) &= \sum_{h \in \mathcal{H}} P(l_y = 1 | h) P(h|X) \\
&=  \sum_{h \in \mathcal{H}} (\alpha \cdot \mathbbm{1}(y \in h) + (1-\alpha) \cdot \beta) P(h|X)\\
&\approx  \sum_{h \in \hat{\mathcal{H}}} (\alpha \cdot \mathbbm{1}(y \in h) + (1-\alpha) \cdot \beta) \hat{P}(h|X)
\end{split}
\end{equation}
where $\mathcal{H}$ is the hypothesis space under the grammar $G$. We also implement two likelihood free parameters $\alpha$ and $\beta$. To account for possible response noise in our collected generalization judgements, we fit a lapse rate $(1-\alpha)$, which determines the probability that a response was made at random. In the case of a lapse trial, we also represent a baseline preference for answering \emph{Yes} ($l_y = 1$) with parameter $\beta$.

Exactly computing the posterior predictive quantity in Eq.~\ref{response1}, requires iterating through all hypotheses in the hypothesis space defined by $G$. Since our grammar $G$ defines an infinite space $\mathcal{H}$ of hypothesized of expressions, we approximate the infinite hypothesis space with a finite set of hypotheses. We construct this finite hypothesis space $\hat{\mathcal{H}} \sim \mathcal{H}$ as follow: we first fix all grammar production probabilities to their default uniform values, and then for each trial $t$ and its set of observed exemplars $X_t$, we estimate a posterior distribution over hypotheses using a Markov chain Monte Carlo (MCMC) inference procedure implemented in the LOTlib3 software package \cite{piantadosi2014lotlib}. Specifically, we run 3 MCMC chains of 100,000 steps of a tree-regeneration Markov chain Monte Carlo (MCMC) procedure \cite{goodman_rational_2008} on each set of exemplars. We then store the top 200 unique hypotheses for each set of exemplars, forming a set that encompasses a large number of hypotheses that are high-probability at some point throughout the experiment. Across all trial types, we obtain a finite space of 5,254 unique hypotheses $\hat{\mathcal{H}}$, and $\hat{\mathcal{H}}$ is used in all subsequent data analyses. Finally, we re-normalize the posterior scores for each $h\in \hat{\mathcal{H}} $ to form a proper posterior distribution $\hat{P}(h\in \hat{\mathcal{H}} |X_t) \propto P(h|X_t)$ for each trial $t$. 

After obtaining a viable hypothesis space $\hat{\mathcal{H}}$, we would like to find the set of grammar production probabilities that most likely generated the observed human categorization judgements. That is, given the set of labels $L$ participants extend to the set of test items $Y$, we are interested in finding the set of grammar parameters $\vec{\theta}$, and likelihood parameters $\alpha, \beta$ such that the (log) likelihood of the behavioral data $P(L | X, Y; \hat{\mathcal{H}})$ is maximized. We also include two temperature parameters that each controls the strength of the prior in Eq.~\ref{prior} and the likelihood in Eq.~\ref{ll}. Together, we optimize for $\argmax_{\bm{\vec{\theta}} }  P(L | X, Y; \hat{\mathcal{H}}, \bm{\vec{\theta}} )$, where $\bm{\vec{\theta}} = \{\vec{\theta}, \alpha, \beta, T_p, T_l\}$. Details of the fitting procedure are described in Appendix \ref{fitting_procedure}.

\subsection{Alternative models}

We compare the full Bayesian program induction model with two lesioned versions, each with parts of the grammar ablated. We also compare with variants of an exemplar model \cite{nosofsky_attention_1986} which, while very successful in modeling human categorization behavior, are not explicitly compositional.

\subsubsection{Bayesian - No defining part} \label{bayes_no_dp}
In this lesioned model (Bayesian no-DP), the nonterminal $DP$ in Fig.~\ref{fullgrammar} and its downstream options are completely turned off, which in turn eliminates all hypotheses that contained one or more defining parts. This means that any concept of the type \textit{a dax is any alien figure that has part $p$ in it} or \textit{a blicket is anything made up of only part $p_1$  or part $p_2$} become unavailable as possible hypotheses in this version of the Bayesian model. The inability to form a defining-part-based concept makes it challenging to grasp concepts like \emph{vehicles} in Fig.~\ref{intro}, which can be defined by objects that have (at least one) wheel(s).

\subsubsection{Bayesian - No variable binding} \label{bayes_no_var}
In this lesioned model (Bayesian no-Var), the ability to bind variables is turned off by disabling the nonterminal $VAR$ in Fig.~\ref{fullgrammar}. All hypotheses that maps a set of parts to a function are unavailable in this model, making any fully abstract pattern difficult to represent. For example, concepts like \emph{a pair of $x$} in Fig.~\ref{intro} which requires a set of variable parts to be duplicated would be a challenging concept for this model without the notion of variable binding. \\

\subsubsection{Exemplar models} \label{exemplar_models}
We evaluate variants of the Generalized Context Model (GCM) \cite{nosofsky_attention_1986} for modeling categorization judgements.
The probability of extending a category label $l_y$ to a new stimulus $y$ is based on its similarity to the training examples $X$:
$$P(l_y = 1 | X) \propto \sum_{i}^{k} \exp(-\sum_{j}^{m} w_j \cdot d_j(y, x_i))$$
where $d_j$ are a set of distance functions (possibly operating over different sets of features) and $w_j$ are the corresponding weight parameters. Following \citeA{overlan_learning_2017}, we convert the raw similarity measures into pseudo probability scores in the range of $[0, 1]$ by normalizing against the maximum similarity over all test items of the trial $t$.

\textbf{Pixel-GCM.} Based on the raw pixel images of the alien figures, we use a deep convolutional neural net (CNN) to extract features of our visual stimuli. A pre-trained 50-layer ResNet (He et al., 2015) is used to encode all images into vectorized representations. There is only one distance function and weight, based on the cosine distance between two feature vectors.

\textbf{String-GCM.} Based on string representations of the alien figures, we use a weighted Levenshtein distance to measure the similarity between exemplars. For every image, its string format is a concatenation of 3 substrings that separately encode 1) shape primitive types, 2) attachment configurations, and 3) overall orientation. For example, an alien figure consisting of two primitives $p_1$ and $p_2$ connected according to their $1^{st}$ allowable configuration and rotated to $180^{\circ}$ can represented in the string format as ``$(p_1p_2)+1+180$''. We fit different weight parameters $w_1$, $w_2$, and $w_3$ for each type of substring, and thus the overall distance $\sum_j^m w_j \cdot d_j(y, x_i)$ is a weighted average of the Levenshtein distances between each pair of corresponding substrings.

\subsection{Results}

\begin{figure}[p]
\begin{center} 
  \includegraphics[width=1.0\linewidth]{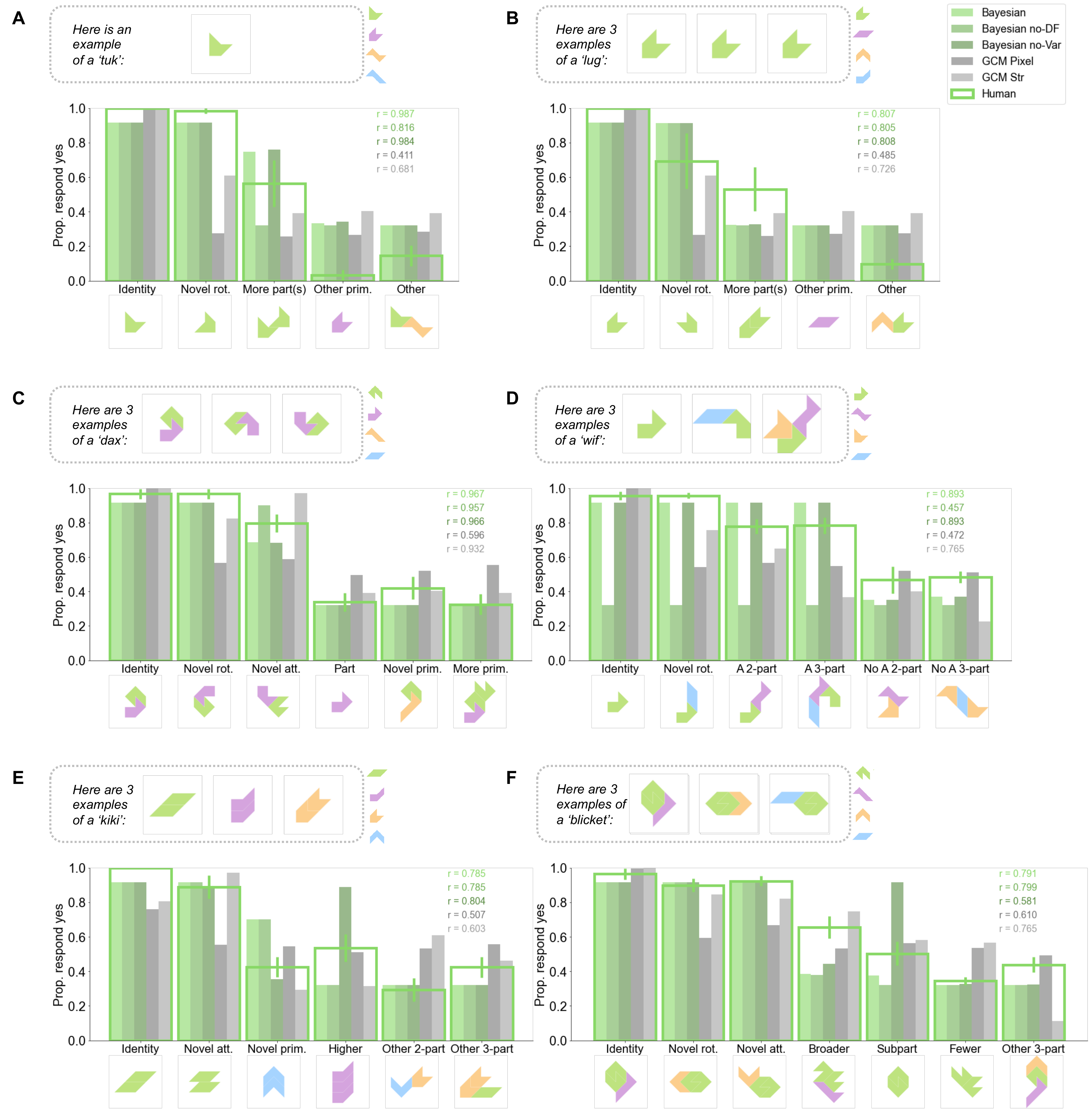}
  \caption{\textbf{Categorization results.} Human behavioral data and model predictions on 6 of the trial types tested in the categorization experiments. The set of training examples is shown at the top of each panel, with the set of 4 candidate shape primitives shown on the right; Test items are categorized into different types of novelty, with examples shown at the bottom. The correlations between human judgments and model predictions per model per trial are indicated in each panel. (A)\&(B) \emph{Identity} test items are identical to one of the examples; \emph{Novel rotation} items are rotated copies of one of the exemplars; \emph{More part(s)} items are more complex compositions of the observed primitives; \emph{Other primitive} are items reflecting novel single shape primitives; \emph{Other} items are conceptually inconsistent with training examples. (C) \emph{Novel attachment} items are new configurations of parts in examples; \emph{Part} test items are parts that appeared in one of the examples; \emph{Novel primitive} items contain unseen parts. (D) Let $A$ be the defining primitive (here, green). \emph{A 2-part} items are two-primitive configurations that contain $A$, \emph{No A 2-part} do not contain $A$; \emph{A 3-part} items are three-primitive configurations that contain $A$, \emph{No A 3-part} do not contain $A$. (E) \emph{Higher (level)} items are configurations with one of the training examples as a subpart; \emph{Other 2-part} \& \emph{other 3-part} are 2-part and 3-part items that do not reflect suggested abstract patterns. (F) \emph{Broader} items are samples from a wider concept for which the set of possible extensions is a superset of the concept of interest; \emph{Subpart} items are the subpart common to all exemplars shown without attachment to some other primitive; \emph{Fewer} items have fewer number of parts than the exemplars.}
  
\label{cat_res}
\end{center} 
\end{figure}

Figure \ref{cat_res} shows human categorization judgments and model predictions on a set of example trial types tested in Experiment 1. The full set of results can be found in Appendix Fig.~\ref{scatters}, which summarizes the relationship between human categorization decisions and model predictions for every trial type and model. Overall, we find that the full Bayesian model provides a strong account of human categorization decisions, achieving a mean correlation of $r=0.901$ across all trial types studied in both experiments. We observe that the Bayesian model consistently assigns high probabilities to the test item that human participants found most likely, and produces graded predictions that tracked people's willingness to extend the concept. 

The alternative models do not perform as well. For the lesioned models, the Bayesian no-DP has an average correlation of $r=0.810$ and the Bayesian no-Var model has an average correlation of $r=0.835$ (Appendix Fig.~\ref{scatters}). As to be expected, the reductions in correlations between human judgments and model predictions for these two models are mainly driven by a subset of trial types that test for concepts that specifically require the notion of defining parts or variable binding. For example, the set of exemplars Fig.~\ref{cat_res}D share an obvious defining part (shown in green) and the Bayesian no-DP model struggles to provide good predictions for all test items. On another trial where there is a subpart common to all exemplars along with some variable part (Fig.~\ref{cat_res}F), the Bayesian no-Var model is able to identify the common subpart but fails to assign higher probabilities to items that also contain a variable part than items that reflect only the subpart.

% \begin{table}
% \begin{center} 
% \caption{Fitted parameters values.} 
% \label{table} 
% % \vskip 0.12in
% \begin{tabular}{ll} 
% \hline
% Type    &  Probability \\
% \hline
% Orientation invariance    &   0.999 \\
% Configuration invariance  &   0.725 \\
% $\alpha$ (1-lapse rate) &   0.839 \\
% $\beta$ (base rate for responding `yes')  &   0.714 \\
% \hline
% \end{tabular} 
% \end{center} 
% \end{table}

\afterpage{\clearpage}
% \begin{figure}[ht!]

The two GCM variants also do not perform at the level of the full Bayesian model. The pixel-GCM has an average correlation of $r=0.577$ and the string-GCM has $r=0.813$. In the case of the pixel-GCM, the model responds strongly to the identity matches, but unlike people, it does not clearly distinguish between the other types of generalization (Fig.~\ref{cat_res}). The pre-trained CNN seemingly fails to perceive the stimuli in terms of their underlying parts and relations, at least without further fine-tuning. The string-GCM is a reasonably good account of the trial types with example figures sharing common parts, but struggles with additional configuration constraints (e.g., Fig.~\ref{cat_res}C). This is unsurprising since the string format precisely encodes which shape primitives are present in each alien figure, but has a less flexible representation of spatial relations. The string-GCM also struggles with more abstract rules that extend to unseen primitives or contain configurations of primitives not previously observed (e.g., Fig.~\ref{cat_res}D\&E). Both variants of GCM do not demonstrate any sensitivity to size principle as such sampling assumption was uniquely built into the likelihood function of the Bayesian model (Fig.~\ref{cat_res}A\&B).

\subsubsection{Fitted parameter values and inductive biases}

The Bayesian model's probabilistic grammar is designed such that its free parameters (probabilities associated with nonterminal symbols) represent psychologically meaningful choices. For example, when sampling a visual concept from the grammar-based prior (see Fig.~\ref{fullgrammar} for full specification), the nonterminal $ATTACH\_SP$ expands in two possible ways (governed by a weighted coin flip): with probability $p_{RI}$ the concept will be orientation invariant, i.e. the set of possible tokens generated by the concept program contains all rotations of a particular configuration; with probability $1-p_{RI}$ the concept will be orientation selectivity such that all possible tokens reflect only one particular orientation. The maximum-a-posteriori (MAP) values of the 8 fitted grammar parameters are reported in Fig.~\ref{params}. Fitted values of the set of grammar parameters reveal a suite of inductive biases people brought to bear when performing this visual concept learning task. For instance, the probability of a rotation invariant concept is near ceiling for the categorization task ($p_{RI}=0.999$), suggesting that our participants had a very strong preference for orientation invariance when judging unnamed alien figures (e.g., Fig.~\ref{cat_res}A). The orientation invariance bias persists even when an increased number of independent exemplars repeatedly show the same orientation, strongly suggestive of orientation selectivity (e.g., Fig.~\ref{cat_res}B). People may have been influenced by their experience with named objects in the real world, which are usually orientation invariant. Participants are also biased towards concepts that do not require fixed configurations of parts, as evident by the similarly high value of the parameter $p_{AI}$. This is exemplified by their willingness to generalize to novel configurations, even when all examples share the same configuration (e.g., Fig.~\ref{cat_res}C).

\subsection{Experiment 1 discussion}

In Experiment 1, we investigate how people can learn a compositional visual concept from just a few examples and then categorize novel examples. The visual concepts are designed to instantiate a variety of different qualitative ways parts can combine, including parts with fixed attachment either with (Fig.~\ref{cat_res}A) or without rotation of the whole figure (Fig.~\ref{cat_res}C), as well as concepts with a defining part (Fig.~\ref{cat_res}B) or defining multi-part motif (Fig.~\ref{cat_res}D) with otherwise variable structure. The Bayesian program induction is able to best match human categorization judgments in comparison to alternative models, demonstrating how a single computational approach can account for a variety of part-based compositional generalization. 

We observe that, even with a very limited number of examples, participants consistently made meaningful generalizations guided by a set of strong assumptions about visual concepts, such as the preference for rotation and attachment invariance. To better understand the types of inductive biases at work, we follow up with an experiment focusing on generating new examples. Participants are asked to generate novel examples of a class of alien figures, after studying a few examples from that class. Our aim is to use generation as a powerful additional window \cite{Ward1994,lake_human-level_2015} into what participants considered as representative for each learned category. Moreover, a complete computational model of concept learning must account for behaviors, like generation, that go beyond classification tasks \shortcite{Markman2003}.

\section{Experiment 2: Few-shot generation of compositional visual concepts}

\subsection{Behavioral experiments}
The few-shot generation task is a modification of the previous categorization experiments, in which the participants studied a set of exemplars and then generated a novel figure belonging to that concept. Details are described below. 

\subsubsection{The generation task} \label{sec:gen_exp}
 
The stimuli used were alien figures identical to the ones that appeared in the categorization experiments. We combined the trial types tested in Experiments 1a and 1b, and formed a set of 31 trials in total which we tested in a single experiment with the generation interface. The order of the 31 trial types was shuffled, and each participant saw a primitive assignment randomly sampled from the 5 possible sets of primitives per trial.

An example trial is shown in Fig.~\ref{procedure} A\&C. Similar to the previous experiments, participants took part in an ``alien figure generation game". In the current task, upon observing a small number of named exemplars, their job as a research assistant was to help generate possible alien figures that belong to the same category as the observed exemplars.

The familiarization procedure was identical to the categorization task. After studying the set of exemplars for each trial, participants entered a test stage in which they used a generation interface to construct an alien figure with the same name. The interface allowed participants to select and drag primitive pieces onto a dynamic digital canvas. The primitive pieces can be freely rotated and connected to other pieces via any of the available sides (attachments that led to overlapping pieces were disallowed). Participants could also fuse pieces together and rotate the fused product as a whole, as well as un-fuse attached pieces and remove any unwanted parts from the canvas. Once they are satisfied with the current composition, participants submitted the final alien figure generation as it existed on the canvas. 

\subsubsection{Participants} 

Participants ($N=135$) were evaluated online and recruited through Amazon's Mechanical Turk. To encourage novel generations (defined as generations that are not exact copies of one of the shown examples), we randomly assigned participants into two groups, each receiving one of two versions of instruction that differed in terms of the emphasis on the expected novelty of generations. Out of all participants, 61 participants were explicitly instructed to generate alien figures that are different from the exemplars, while 74 participants received the version of instruction that did not specifically require the generations to be novel. We did not find a significant difference in the average percentage of novel generations between the two groups ($63.6\%$ for the first group and $64.8\%$ for the second), and we pooled the data from the two instruction groups in all subsequent analyses.

All participants finished the task within an hour and were paid \$5.00 at the completion of the experiment. As an attention check, participants completed a set of quiz questions about the generation game interface. Participants were given five chances to answer the quiz questions correctly, although 5 people were unable to and thus excluded.

We observed two distinct types of strategies that participants adopted across trials: one strategy involved consistently copying one of the provided exemplars, while the other strategy involved consistently generating a novel alien figure not present in the exemplar set. We divided the participants into two groups according to their adopted strategy, with one group containing 42 participants that only copied, and another group containing 88 participants that produced novel generations. As we are interested in modeling generalization behaviors, generated alien figures from the second group of participants were used in our reported analyses.

\subsection{Bayesian program induction}
The Bayesian program induction model (Section \ref{sec_BPI}) can both classify and generate novel examples. For trial $t$, consider figures $Y_t$ generated by participants in response to the set of exemplars $X_t$. To generate each new example $y \in Y_t$, we can sample from the posterior predictive distribution $P(y|X_t; \vec{\theta}, \alpha)$:
 \begin{equation} \label{predicative}
\begin{split}
P(y|X_t; \vec{\theta}, \alpha) = \sum_{h\in H} \underbrace{P(y|h; \alpha)}_{i}P(h|X_t; \vec{\theta})
\end{split}
\end{equation}
where the term ($i$) on the right-hand-size of Eq.~\ref{predicative} is the likelihood of a new example if $h$ is true. It has the form 

$$ P(y|h; \alpha) = \alpha \cdot \mathbbm{1}(y \in h)\cdot \frac{1}{|h|} + (1-\alpha) \cdot P^{0}(y),$$ where $\mathbbm{1}(\cdot)$ indicates whether $y$ is a valid token of $h$, and $|h|$ is the size of the given hypothesis $h$, the number of all possible tokens under $h$. We again included a lapse rate: the new example is noisily generated by either sampling from all valid tokens of $h$ with probability $\alpha$ or by sampling from the null token distribution (see Appendix \ref{p0}) of tokens $P^{0}(x)$ with probability $1-\alpha$. 

Since both the categorization and generation experiments shared the same set of trial types, we can use the same set of hypotheses $\hat{\mathcal{H}}$ to approximate the infinite hypothesis space  $\mathcal{H}$ considered in the generation task. Once again, we re-normalize the posterior scores for each $h\in \hat{H}$ to form a proper posterior distribution and the response distribution $P(Y_t | X_t; \vec{\theta}, \alpha)$ becomes:

\begin{equation} \label{predicative2}
\begin{split}
P(Y_t|X_t; \vec{\theta}, \alpha) & = \prod_{y\in Y_t} P(y|X_t; \vec{\theta}, \alpha) \\
&= \prod_{y\in Y_t} \sum_{h\in \hat{\mathcal{H}}} P(y|h; \alpha)\hat{P}(h|X_t; \vec{\theta})
\end{split}
\end{equation}

To infer the set of un-observable model parameters using the human generation data, we are interested in finding the set of parameter values $\bm{\vec{\theta}}$ that maximizes the (log) likelihood $P(Y | X; \bm{\vec{\theta}})$ of all participant generations $Y$, upon observing exemplars $X$. Again, we include two temperature parameters that control the strengths of the prior and likelihood respectively, hence $\bm{\vec{\theta}} = \{\vec{\theta}, \alpha, T_p, T_l\}$. The subsequent parameter fitting procedure is identical to that of Experiment 1, and reported in Appendix \ref{fitting_procedure}.

Instead of refitting the set of grammar production probabilities, we also evaluate performance of the Bayesian model with the set of MAP values for $\vec{\theta}$ directly transferred from Experiment 1. We expect the Bayesian model with transferred parameter values to perform at a comparable level to the Bayesian model refitted to the generation dataset, if participants' assumptions about the alien figure concepts remain consistent across different tasks.

%and we calculated the overall likelihood of each set of parameters by taking the product of the likelihood score of each human generated example $y_{t,i}$ over all $T$ trials:

% \begin{equation} \label{logscore}
% \begin{split}
% \argmax_{\bm{\vec{\theta}}}  P(Y | X; \bm{\vec{\theta}}) = \argmax_{\bm{\vec{\theta}}}  \prod_t^T \prod_i^n P(y_{t,i} | X_t; \bm{\vec{\theta}})
% \end{split}
% \end{equation}
% where $n$ is the number of participant generation collected for each trial.

% We also included two more free parameters in $\bm{\vec{\theta}}$, a prior temperature $T_p$, and a likelihood temperature $T_l$ which control the strength of the prior in eq.\ref{prior} and likelihood in eq.\ref{ll} by raising them to the $1/T$th power, respectively. We re-normalized the temperature-adjusted prior distribution to be $\hat{P}(h;\vec{\theta}, T_p) \propto P(h; \vec{\theta})^{1/T_p}$, and we subsequently re-normalized the posterior distribution after likelihood temperature adjustment to be $\hat{P}(h|X_t; \vec{\theta}, T_p, T_l) \propto  \hat{P}(h;\vec{\theta}, T_p)P(X_{t}|h)^{1/T_l} \propto P(h|X_t; \vec{\theta})$ .  The data likelihood score hence becomes:

% \begin{equation}\label{logscore2}
% \begin{split}
% P(Y | X; \bm{\vec{\theta}}) &= \prod_t^T \prod_i^n P(y_{t,i} | X_t; \bm{\vec{\theta}})\\
% &= \prod_t^T \prod_i^n \sum_{h\in \hat{\mathcal{H}}} P(y_{t,i}|h; \alpha)\hat{P}(h|X^E_t; \vec{\theta}, T_p, T_l)
% \end{split}
% \end{equation}

\subsubsection{Alternative models}

We again compare the full Bayesian program induction model with two lesioned variants, Bayesian no-DP (Section \ref{bayes_no_dp}) and Bayesian no-Var (Section \ref{bayes_no_var}). We also compare to a variant, Bayesian Exp. 1 fit, that copies over the parameter fits from the Experiment 1 categorization task.

We also compare the Bayesian model with a generative variant of the string-GCM  (Section \ref{exemplar_models}). We convert this exemplar model into a generative model by enumerating all possible tokens (in the string format), defining the probability of a generated exemplar $y$ as:
$$P(y | X_t) = \frac{\sum_{i}^{k} \exp(-\sum_{j}^{m} w_j \cdot d_j(y, x_i))}{\sum_{y \in S} \sum_{i}^{k} \exp(-\sum_{j}^{m} w_j \cdot d_j(y, x_i))},$$
where the set of provided exemplars is $X_t$ and $S$ is the set of all possible token strings.

\subsection{Results}

Table~\ref{tab:gen_results} provides a summary of how the full Bayesian program induction model compares to alternatives. Using mean log-likelihood per human generated token, the full Bayesian model shows the strongest overall performance. The Bayesian model with parameters from Experiment 1 is the next strongest performer, although the drop in performance suggests differences between the categorization and generation tasks (which are discussed below). The next best models are the lesioned variants Bayesian no-DF and Bayesian no-Var. The lowest performing model is the string-GCM model. Paired t-tests between the full Bayesian model and each of the alternatives, with per-token log-likelihood values as observations, confirm the differences (details in Table~\ref{tab:gen_results}). 
% log-likelihood difference $\ell(\theta) - \ell(\theta_0)$ between each alternative model, $\theta$, and the refitted full Bayesian model $\theta_0$ show a significantly better performance by the Bayesian model in comparison to the Bayesian no-DP model [t(1708) = -8.780, p < 0.001], the Bayesian no-Var model [t(1708) = -8.117, p < 0.001], as well as the string-GCM [t(1708) = -49.369, p < 0.001]. Same analysis also indicates that the Bayesian model suffers a decline in performance with transferred parameters [t(1708) = -11.387, p < 0.001], suggesting a possible divergence in the set of assumptions participants make about the visual concepts across different tasks.

\begin{table}[ht]
    \centering
    \begin{tabular}{lccc}
        \toprule
        Model & log-likelihood & t-statistic (p-value) \\
        \midrule

        Bayesian & \textbf{ -5.177 } &  - \\
        Bayesian (Exp. 1 fit) & -5.404  &  -11.387 (0.000)\\
        Bayesian no-DF  &  -6.256          &  -8.780 (0.000) \\
        Bayesian no-Var &  -5.760          &  -8.117 (0.000) \\
        String-GCM      &  -10.354         &  -49.369 (0.000)  \\
        \bottomrule
    \end{tabular}
    \caption{\textbf{Goodness of fit for predicting human generated examples.} For each model, the average log-likelihood per human generated token is reported in the first column. Paired t-tests compare the full Bayesian model to each alternative (with 1708 degrees of freedom). The resulting t-stats and p-values are shown.}
    \label{tab:gen_results}
\end{table}

\begin{figure}[ht!]
\begin{center} 
  \includegraphics[width=0.9\linewidth]{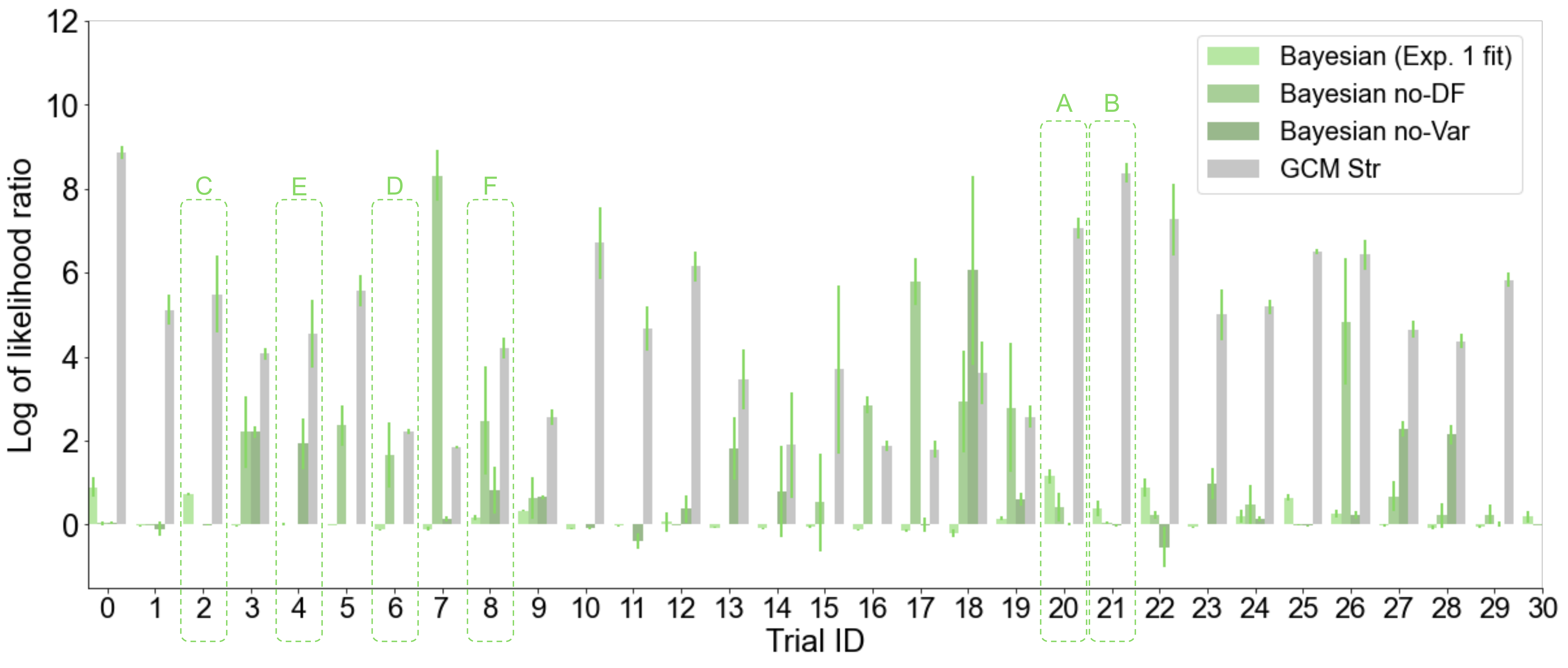}
  \caption{\textbf{Log of likelihood ratios per trial type.} Mean per-token (log) likelihood ratio between the full Bayesian program induction model and each alternative model per trial type averaged over different random assignments of part primitives. A bar in the positive direction suggests a better log-likelihood score predicted by the full Bayesian model as opposed to the alternative model. Trial types corresponding to the examples in Fig.~\ref{gen_res} are indicated.}
  
\label{log_ratios}
\end{center} 
\end{figure}

\begin{figure}[p]
\begin{center} 
  \includegraphics[width=0.98\linewidth]{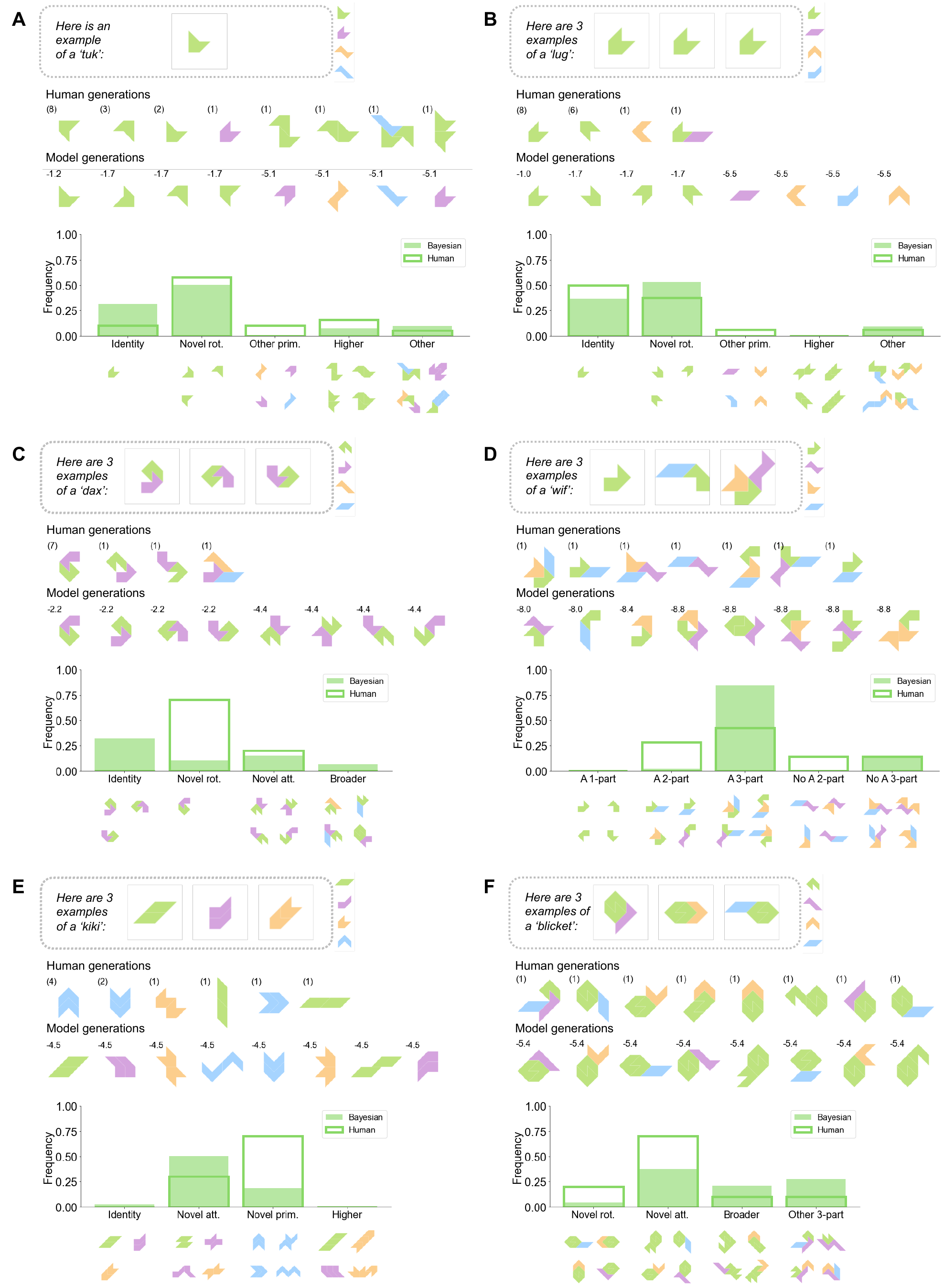}
  \caption{\textbf{Generation results.} For the same 6 trial types in Fig.~\ref{cat_res}, the most frequent human generations (with frequency in upper left) and most likely Bayesian model generations (with log-likelihood in upper left) are compared. Below these individual examples, a bar graph categorizes the human/model samples by the type of novelty they represent (see Fig.~\ref{cat_res} for description of each type).}
  
\label{gen_res}
\end{center} 
\end{figure}

Fig.~\ref{log_ratios} shows mean difference in log-likelihood per trial, $\ell(\theta) - \ell(\theta_0)$ for full model $\theta$ and alternative $\theta_0$, such that positive values mean the full Bayesian model is favored. To highlight several key trials, Fig.~\ref{gen_res} shows human generations alongside Bayesian model samples. The string-GCM consistently produces the poorest log-likelihoods across all trial types. All variants of the Bayesian model, with their built-in notions of invariance, show preference to produce samples that generalize outside of observed orientations (Fig.~\ref{gen_res}A\&B) and part attachments (Fig.~\ref{gen_res}C). Consistent with findings from Experiment 1, the two lesioned Bayesian models mainly deviate from the full version on a subset of the trials that require having defining parts or variable manipulations. Again, when all exemplars in Fig.~\ref{gen_res}D share the green defining part, the Bayesian no-DF model struggles, and when there is an abstract pattern fulfilled by variable parts (Fig.~\ref{gen_res}E) the Bayesian no-Var model struggles. When there is both a defining common subpart and a variable part across all observed examples (Fig.~\ref{gen_res}F), both lesioned models fall short in comparison to the full model. 

The refitted full Bayesian model not only outperforms alternative models in terms of log-likelihoods, it is also can also generate compelling new examples that resemble human generations. For example, the most frequently generated examples in Fig.~\ref{gen_res}E correctly capture the abstract pattern. The most frequent generations in  Fig.~\ref{gen_res}F share the common subpart with the provided set of exemplars. However, we also notice a number of qualitative discrepancies between human generations and model produced samples. For example in Fig.~\ref{gen_res}C, participants overwhelmingly prefer the token with a novel orientation, whereas the model assigns equal probability to tokens reflecting the same composition at all orientations. In Fig.~\ref{gen_res}E, participants demonstrate a strong preference for the novel (blue) primitive, while the Bayesian model shows no such preference. We examine these phenomena more closely in Section \ref{sec_inductive_biases}. Additionally, both humans and the Bayesian model are able to identify the green defining part in Fig.~\ref{gen_res}D, as indicated by the common green part in most human and model generated tokens. However, when participants are asked to generate a new token with the green defining part, they generate 2-part tokens at a frequency only slightly lower than that of 3-part tokens. The Bayesian model, on the other hand, produces mostly 3-part samples due to the uniform nature of its likelihood function, as there are many more 3-part tokens than 2-part tokens within the same concept.

% \begin{wrapfigure}{r}{0.7\textwidth}
%     \centering
%     \includegraphics[width=0.69\textwidth]{images/top5acc.pdf}
%     \caption{Top-5 generation accuracy. \todo{moving to supplementary materials} }
%     \label{fig:top5acc}
% \end{wrapfigure}

% Overall, we found that participants were able to make meaningful inferences and construct novel visual forms spanning a wide range of composition types after observing only a small number of positive exemplars. The Bayesian program induction model was also able to make predictions that echo human generations across different trial types, as measured by the log-likelihood scores of human generated alien figures predicted by the model.

%log-likelihood $\ell(\theta) - \ell(\theta_0)$ between the GNS model, $\theta$, and Bayesian model $\theta_0$ [t(336) = 6.197, p < 0.001]

\afterpage{\clearpage}
% \begin{figure}[ht!]

\subsubsection{Inductive biases} \label{sec_inductive_biases}

\begin{figure}[ht!]
\centering
\includegraphics[width=0.99\linewidth]{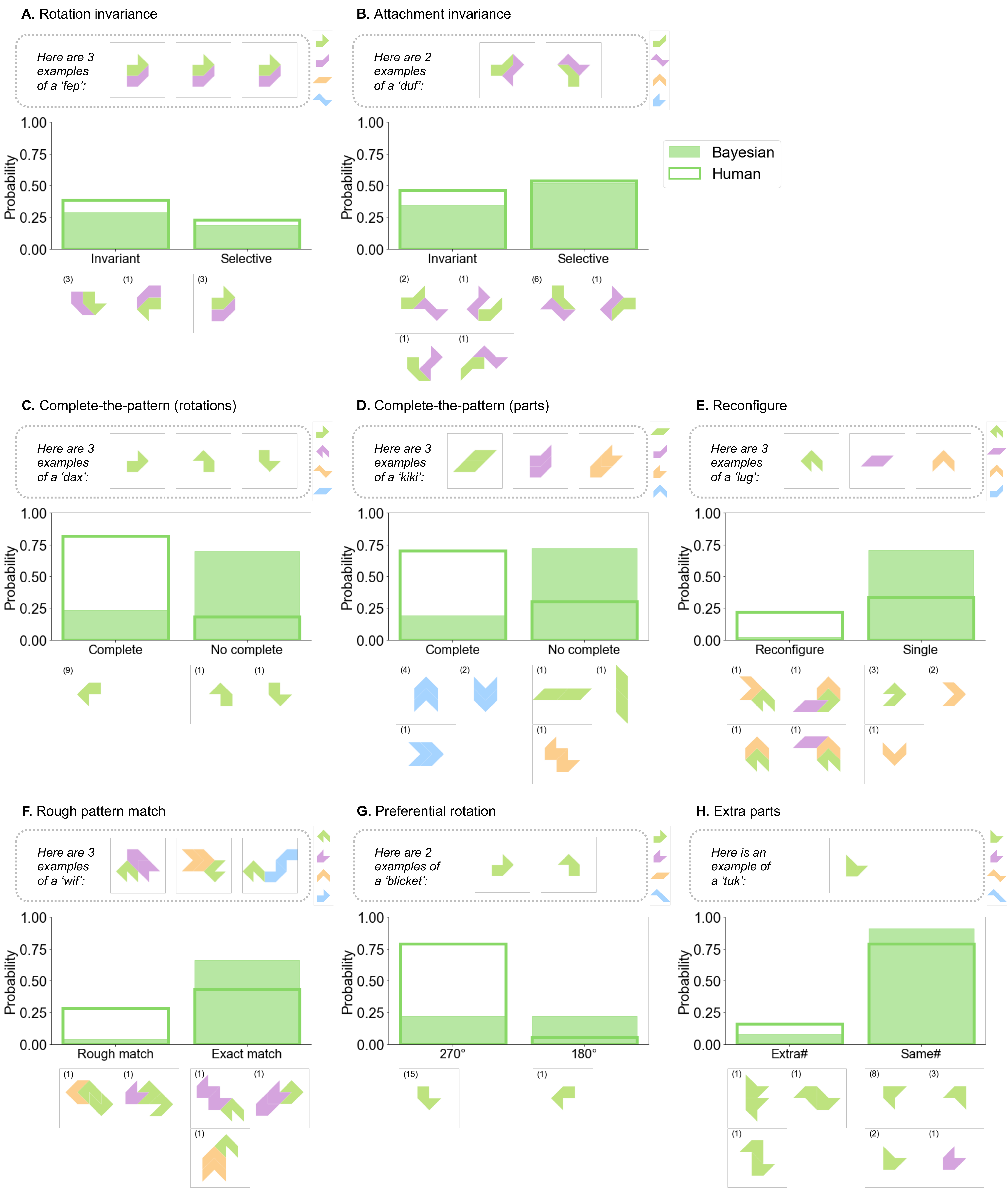}

\caption{{\bf Inductive biases in compositional visual concept generation.} Example trials that demonstrate various human inductive biases. The bar plot in each panel shows the average probabilities predicted by the Bayesian model along with empirical frequencies for generations that follow the bias vs. violate the bias. Examples of human generated tokens for each category are shown on the bottom of each panel along with the associated raw counts of occurrences in the data. }

%(A) On trials where there were repetitions of tokens at identical orientations, participants were likely to generate tokens with the same attachments but at novel orientations (left). The Bayesian model was able to replicate this orientation invariance bias (right). (B) Participants also demonstrated a configuration invariance by generating tokens that with the same set of primitives but attached differently than the exemplars (left), the same inductive bias was well captured by the Bayesian model (right). }
\label{biases}
\end{figure}

In the generation experiment, we observe evidence for a set of inductive biases that appear to guide participants' generalizations, especially in the tasks studied here that have a limited number of provided examples. After tuning free parameter values to the human data, we examine whether the fitted Bayesian model is able to reproduce generation patterns similar to those of human participants. Specifically, for each trial and bias type, we identify the set of all alien figure tokens that are consistent with the given inductive bias, and compare the probability of generating bias tokens predicted by the Bayesian model to the frequency of bias tokens in the behavioral data (Fig.~\ref{biases}). We find that a fitted Bayesian model is successful in capturing some of the human inductive biases, but lacks the mechanisms needed to capture the more subtle statistical patterns of behavior uncovered in the generation task. 

\paragraph{i. Inductive biases accounted for by the Bayesian model.}

The two inductive biases about invariance assumptions are well captured by the Bayesian model, consistent with the findings in Experiment 1.

\textbf{Orientation invariance} is a preference for assigning rotated variants of the same figure to the same concept. This is illustrated in trial Fig.~\ref{biases}A: three provided examples of the same token is suggestive of an orientation selective concept \cite{xu_word_2007}, yet many participants used a novel orientation. The fitted Bayesian model confirms this preference for novel orientations, as evident by the high value of the $p_{RI} = 0.936$ parameter (Fig.~\ref{params}).

\textbf{Attachment invariance} is a preference for assigning all alien figures with the same parts to the same concept (regardless of attachment relation). This is illustrated in trial Fig.~\ref{biases}B: two provided examples utilize the same attachment, yet many participants used a new attachment. The fitted Bayesian model confirms this preference, with the parameter $p_{AI} = 0.584$ (although notably, this preference is not as strong in the categorization data).

\paragraph{ii. Inductive biases not accounted for by the Bayesian model.} Different from the biases above, which had designated parameters in the model that control invariance assumptions, there are a number of distinct behavioral patterns that are beyond the model's current capabilities.

\textbf{Complete-the-pattern} is a preference for generating an exemplar that completes a set along a particular dimension. We observed two variants of this preference when generating a new exemplar: observing exemplars with 3 distinct rotations and choosing the 4th (and last) rotation option (see example in Fig.~\ref{biases}C), or observing exemplars that each use a different primitive and choosing the 4th (and last) primitive option (see example in Fig.~\ref{biases}D). This bias is especially interesting due to the violation of an extremely common Bayesian modeling assumption: independent and identically distributed sampling of data points (Eq.~\ref{ll}). Thus, instead the Bayesian model assigns equal probability to tokens regardless of which orientation is chosen in Fig.~\ref{biases}C and which primitive is duplicated in Fig.~\ref{biases}D.

\textbf{Reconfigure} is a preference for using parts from existing figures to compose more complex figures reconfigure the parts (Fig.~\ref{biases}E). This is an alternative strategy to generate novel tokens that some participants adopted when they are not completing a pattern as defined above. Although the Bayesian model is able to produce more complex samples utilizing familiar parts, they have much lower probabilities than the novel tokens that contain only a single part.

% \begin{figure}[ht!]
% \centering
% \includegraphics[width=0.8\linewidth]{SVC/images/biases2.pdf}

% \caption{{\bf Inductive biases in compositional visual concept generation (2). } (A) A complete-the-pattern bias was observed when the set of exemplars included the same token presented at all but one canonical orientations. Participants overwhelmingly produced a new example oriented at the $4^{th}$ canonical position. (B) A similar complete-the-pattern bias occurred when all exemplars shared the same abstract pattern, using all but one of the available shape primitives. Participant showed a strong preference for the unused shape primitive when composing their own alien figure (left). Samples from the Bayesian model did not show a similar strong tendency to ``complete the pattern". (C) Participants sometimes demonstrated a reconfigure bias, composing more complex alien figures using familiar single tokens as parts.}
% \label{biases2}
% \end{figure}

\paragraph{iii. Other inductive biases} This is a set of more subtle inductive biases with lower occurrences in the human generations. Some are likely shortcuts to ensure generations are distinct from all exemplars (but not necessarily conceptually consistent); others are likely behavioral artifacts of the experimental interface.

\textbf{Rough pattern match} is a partial sensitivity to abstract patterns of parts, although with a characteristic swapping of variables (Fig.~\ref{biases}F). For example, when all exemplars show a ``$x$-$x$-$A$" pattern, some participant-generated alien figures reflect an ``$A$-$A$-$x$" pattern instead. The Bayesian model is unable to account for this type of response.

\textbf{Preferential orientation} is a preference for one choice of rotation over another, without a clear inductive explanation (Fig.~\ref{biases}G). We hypothesize this to be a possible artifact of the experimental interface, as rotating a composed alien figure positioned at $0^{\circ}$ in the game to $270^{\circ}$ requires 1 double-click, but 3 double-clicks are needed for a $180^{\circ}$ rotation. Hence, participants generally favor the option that involves less manual work. The Bayesian model has no inherent preference for one specific orientation over others.

\textbf{Novelty by adding extra parts} is a preference for adding primitives to an existing exemplar to make a novel one. This is observed on highly open-ended trials like Fig.~\ref{biases}H where there are other options to make novel exemplars from a single part.

% \begin{figure}[ht!]
% \centering
% \includegraphics[width=0.8\linewidth]{SVC/images/biases3.pdf}

% \caption{{\bf Inductive biases in compositional visual concept generation (3). } (B) On trials where there are two conceptually consistent options that only differed in terms of their orientations, participants overwhelmingly favored one rotation over the other. This is likely an artifact of the game interface (left). The Bayesian model was not equipped to favor some orientations  over others (right). (C) When the trial was highly open-ended and contained only one example, participants tended to produce a more complex configuration than the exemplar to induce novelty in their generations (left). The Bayesian model did not make similar broad generalizations (right).  }
% \label{biases3}
% \end{figure}

\subsection{Experiment 2 discussion}

Overall, we find that participants are able to make meaningful inferences and construct novel visual forms spanning a wide range of composition types after observing only a small number of positive exemplars. Interestingly, by explicitly asking participants to generate their own examples, we elicit patterns of behavior that diverge from what we observe in human categorization judgements using the same sets of stimuli. For example, the complete-the-pattern bias in Fig.~\ref{biases}C\&D is unique to the generation task. In fact, behavioral data from Experiment 1 show the opposite effect, as indicated by the drop in generalization to logically consistent test items with novel primitives (see a more detailed example in Appendix Fig.~\ref{fig:transfer}). These qualitative behavioral differences also contribute to the distinct MAP values reported in Appendix Fig.\ref{params} when the Bayesian model is fitted separately on the two sets of experimental data; they also help explain the decline in model performance when Experiment 1 grammar parameters are used to describe the human generations in Experiment 2.

In addition to the complete-the-pattern bias, the generative task reveals a richer set of human inductive biases for learning compositional visual concepts. We find that our Bayesian program induction model generates compelling new examples that resemble human generations (Fig.~\ref{gen_res}) and accounts for some of the inductive biases with a single re-write probability parameter in the PCFG prior. However, other behaviors are beyond the model's current capabilities, including violations of the independence assumptions for how exemplars are generated, unusual combinations of parts, and other abstractions not considered as part of the PCFG prior. To better account for these additional behaviors, in the next section, we introduce a hybrid neuro-symbolic program induction model that combines the types of compositional representations used in the Bayesian model with data-driven components for additional modeling power.

% %\section{GNS}
% \input{gns.tex}

\section{Experiment 3: Capturing additional behavioral structure with generative neuro-symbolic modeling}

Symbolic probabilistic models like ours provide an elegant and interpretable account of human behavior; however, these models make simplifying and rigid parametric assumptions, and as result, they often leave portions of the data unexplained.
For example, the Bayesian program induction model assumes that all constituent tokens $x_i$ of a hypothesis $h$ are sampled with equal probability (Eq.~\ref{ll}). 
This assumption appears at odds with humans, who at times exhibit a preference for certain tokens over others within a particular grouping (Fig. \ref{biases}C\&D).
Although there may be an ad-hoc rule to explain each behavioral nuance like this, engineering such primitives would involve a considerable effort, and the complexity of the resulting model could quickly grow out of hand.
Alternatively, we could let the data speak for itself by integrating more powerful data-driven modeling components.

In pursuit of a more complete computational account with much of the same structure and interpretablility, we propose to model human concepts of alien figures as neuro-symbolic probabilistic programs.
This paradigm, known as Generative Neuro-Symbolic (GNS) Modeling, was shown to provide an effective framework for understanding another type of compositional visual concept: handwritten letters from different alphabets \cite{Feinman2021}. 
As in the fully-symbolic Bayesian model, the aim of GNS is to infer the best causal generative process for explaining the visual examples. 
Unlike symbolic models, GNS further represents nonparametric statistical relationships between parts in a token, and between tokens in an observation, providing a more flexible model with fewer a priori assumptions.
Moreover, a GNS model can be estimated directly from training data, providing an effective data-driven approach.
An important component of our approach is that we train GNS to mimic the Bayesian program induction model by using the Bayesian model to generate some of its training data, while also including real human data so that GNS can go further to capture additional structure in human behavior.

\subsection{Model Description}

\begin{figure}[ht!]
    \centering
    \includegraphics[width=0.95\hsize]{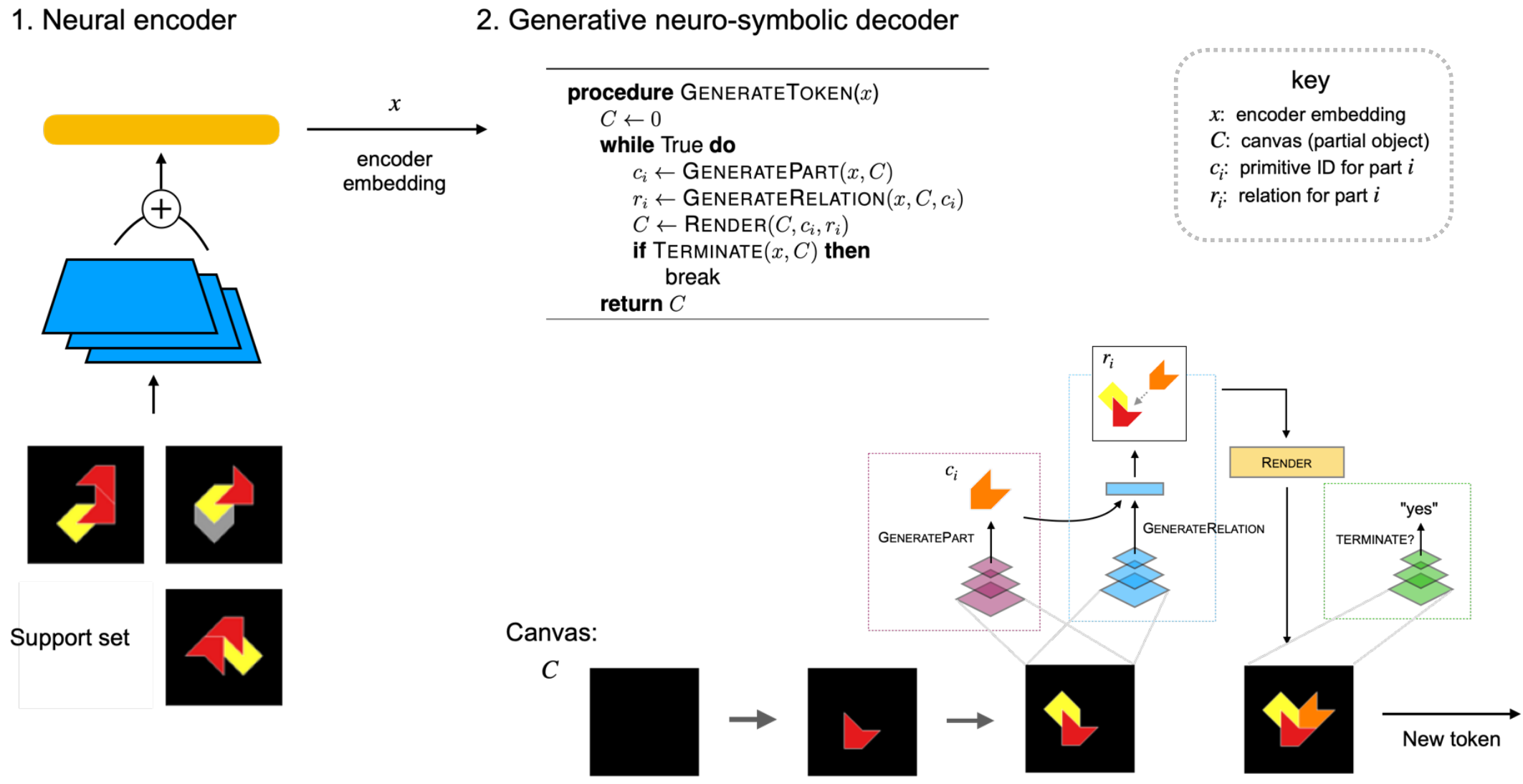}
    \caption{\textbf{Overview of GNS model.} A neural encoder first reads each support example with a convolutional neural network (CNN) and aggregates the resulting vectors into a single, fixed-sized embedding. 
    This encoder embedding is then passed to a GNS decoder--expressed as probabilistic program \texttt{GenerateToken}--that generates new tokens one part at a time, using an image canvas $C$ as memory. 
    At each part iteration $i$, the current canvas $C$ and encoder embedding $x$ are first fed to subroutine \texttt{GeneratePart} which generates the primitive ID $c_i$ of the next part. Next, $C$, $x$ and $c_i$ are passed to subroutine \texttt{GenerateRelation} which samples a relation specification $r_i$ for the part. Finally, a symbolic renderer updates the canvas according to $c_i$ and $r_i$, and subroutine \texttt{Terminate} decides whether to terminate the token.
    }
    \label{fig:gns_model_overview}
\end{figure}

A depiction of the proposed GNS model is given in Fig. \ref{fig:gns_model_overview}. Similar to a previous model of handwritten characters \cite{Feinman2021}, our GNS model of alien figures uses the control flow of a probabilistic program, coupled with an external image memory, to represent the causal process of generating new concepts.
Through repeated calls to subroutines \texttt{GeneratePart} and \texttt{GenerateRelation} the model maintains a representation that is \textit{compositional}, providing and appropriate inductive bias for compositional generalization.
Each of these modular subroutines is expressed as a neural network that generates symbolic outputs conditioned on the current program state (Fig. \ref{fig:gns_subroutines}).
New from prior work, we augment the GNS model with an image encoder to account for the ways that people induce conceptual representations from exemplars in the current behavioral experiment.
With this addition, we can use our GNS model as a proxy to the Bayesian model's posterior predictive distribution (Eq. \ref{predicative}).
Given a set of support exemplars, the encoder first reads each exemplar using a convolutional neural network (CNN) and then aggregates the individual responses to form a single vector embedding of the set. This embedding is passed to the decoder and used to condition a generative model for new tokens.
Both the encoder and the decoder use a coloring scheme for alien figure images that associates each primitive from our primitive bank with a unique RGB color.

\subsubsection{Encoder}
The support encoder (Fig. \ref{fig:gns_model_overview}, left) consists of a convolutional neural network (CNN) backbone and a transformer aggregator. The CNN first reads each exemplar in the support set, represented as an $80 \times 80$ RGB image, and encodes the image to a 256-dimensional vector. The sequence of CNN vectors is then fed to a transformer encoder, which processes the variable-length sequence and outputs an aggregate vector encoding of the set.

\subsubsection{GNS Decoder}
Our GNS decoder (Fig. \ref{fig:gns_model_overview}, right) generates new tokens by sampling a sequence of symbolic primitives $\{\theta, c_{1:\kappa}, r_{1:\kappa},\}$ which together specify a unique instance of an alien figure concept with $\kappa$ parts.
Part assignments $c_i$ convey the category of the $i^{th}$ part, chosen from a dictionary of 9 basic primitive categories, and relations $r_i$ specify how the $i^{th}$ part attaches to previously-generated parts, with $r_1$ assigned to null. Each attachment specification $r_i$ encompasses 3 unique sub-choices: an index $j$ of the previous part onto which the current part $i$ will attach, and a choice of polygon sides $s_j$ and $s_i$ for the previous and current part that will touch at the point of attachment.

\begin{figure}[ht!]
    \centering
    \includegraphics[width=\hsize]{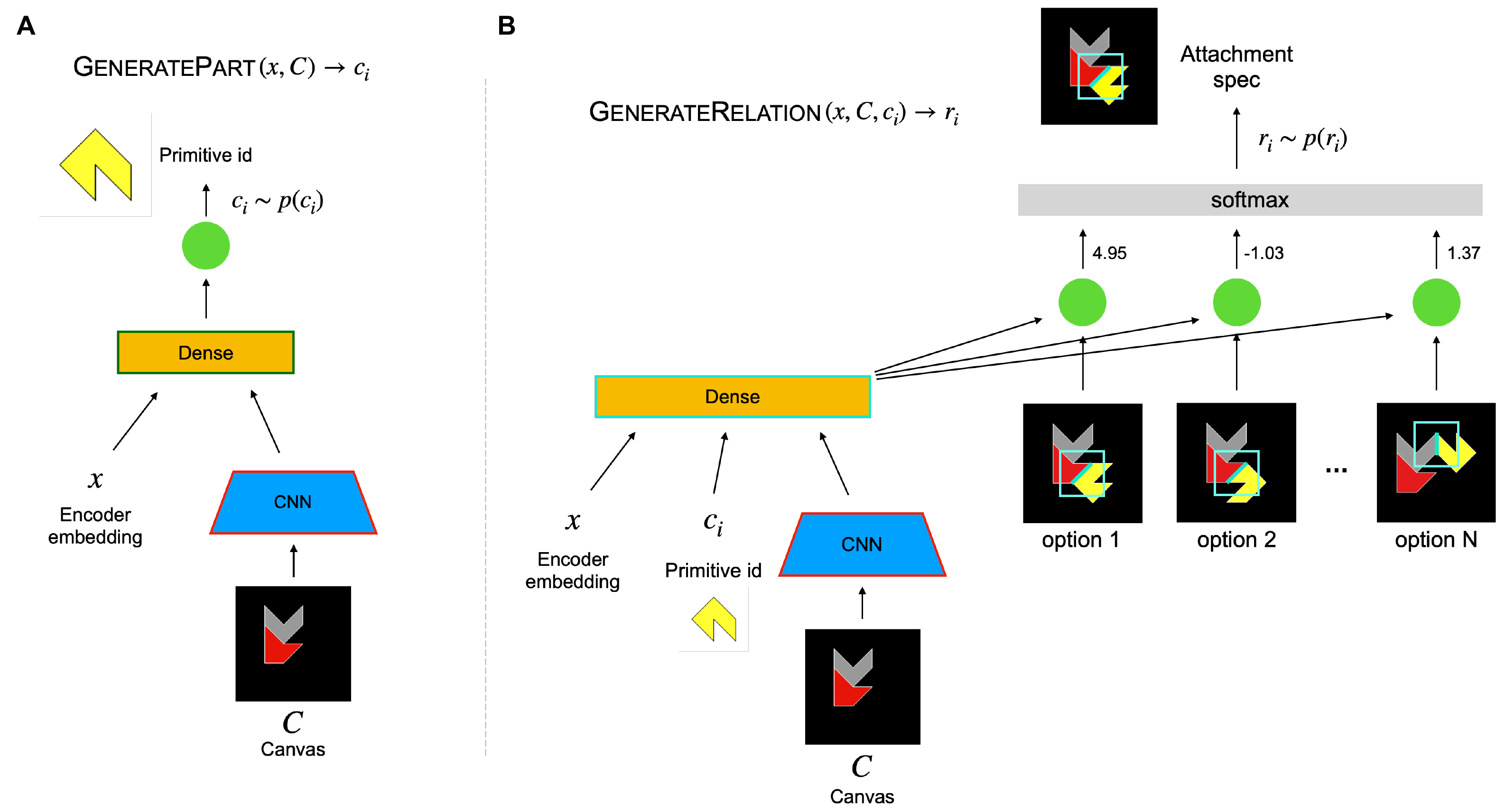}
    \caption{\textbf{GNS Subroutines.} (A) Subroutine \texttt{GeneratePart} first reads the image canvas with a CNN and concatenates the response with encoder embedding $x$. The combined vector is then processed by a dense layer and passed to a softmax prediction head that yields a categorical distribution to sample the next primitive ID $c_i$. (B) Subroutine \texttt{GenerateRelation} similarly reads the canvas with a CNN, this time concatenating with both the encoder embedding $x$ as well as primitive ID $c_i$ from \texttt{GeneratePart}. The combined vector is processed by a dense layer and then passed to a relation prediction head that yields a probability distribution to sample the next relation $r_i$ (see Fig. \ref{fig:gns_relation_prediction_architecture}) for additional details.}
    \label{fig:gns_subroutines}
\end{figure}

The generative process to sample a new token conditioned on support embedding $x$ proceeds as follows. We first initialize an empty image canvas, $C$, that will maintain the state of the sample. Next, we sample a global orientation $\theta$ for the token from subroutine \texttt{GenerateOrientation}. This is an additional neural network module that is used only once at the start of the sample and it selects from 4 discrete orientation choices.
From there, we iteratively sample the next part and next relation from subroutines \texttt{GeneratePart} and \texttt{GenerateRelation} until a termination is reached.
Each of these sample steps conditions on the support, as well as the current partial-object, by reading $x$ and $C$ as neural network inputs. This design enables the model to capture complex correlations that permeate through multiple parts of an object, or that connect a new object to support examples.
At the end of each iteration, we update our canvas $C$ with the latest partial-object using a symbolic image renderer and pass the new canvas to subroutine \texttt{Terminate}, a neural network that decides whether to terminate the object or continue with another part.

The architectures of the neural networks for \texttt{GeneratePart} and \texttt{GenerateRelation} are depicted in Fig. \ref{fig:gns_subroutines}. 
In \texttt{GeneratePart}, a CNN embeds the current image canvas to a vector and concatenates it with the encoder embedding. The combined vector is then processed by a fully-connected (dense) layer, and a softmax layer then predicts a categorical distribution for the primitive ID of the next part.
In \texttt{GenerateRelation}, a CNN similarly encodes the image canvas, this time concatenating the resulting vector with both the encoder embedding as well as a discrete embedding of the primitive ID chosen in the previous step. 
The concatenated vector is then processed by a dense layer and fed to an attention-style prediction head. Using this input and an attention-style weighting scheme, the prediction head outputs a distribution over discrete choices for how and where the new part will attach to previous ones in the canvas (Fig. \ref{fig:gns_relation_prediction_architecture}).

\subsection{Training with meta-learning}
\label{sec:training_with_metalearning}

\begin{wrapfigure}{r}{0.6\textwidth}
    \centering
    \includegraphics[width=0.59\textwidth]{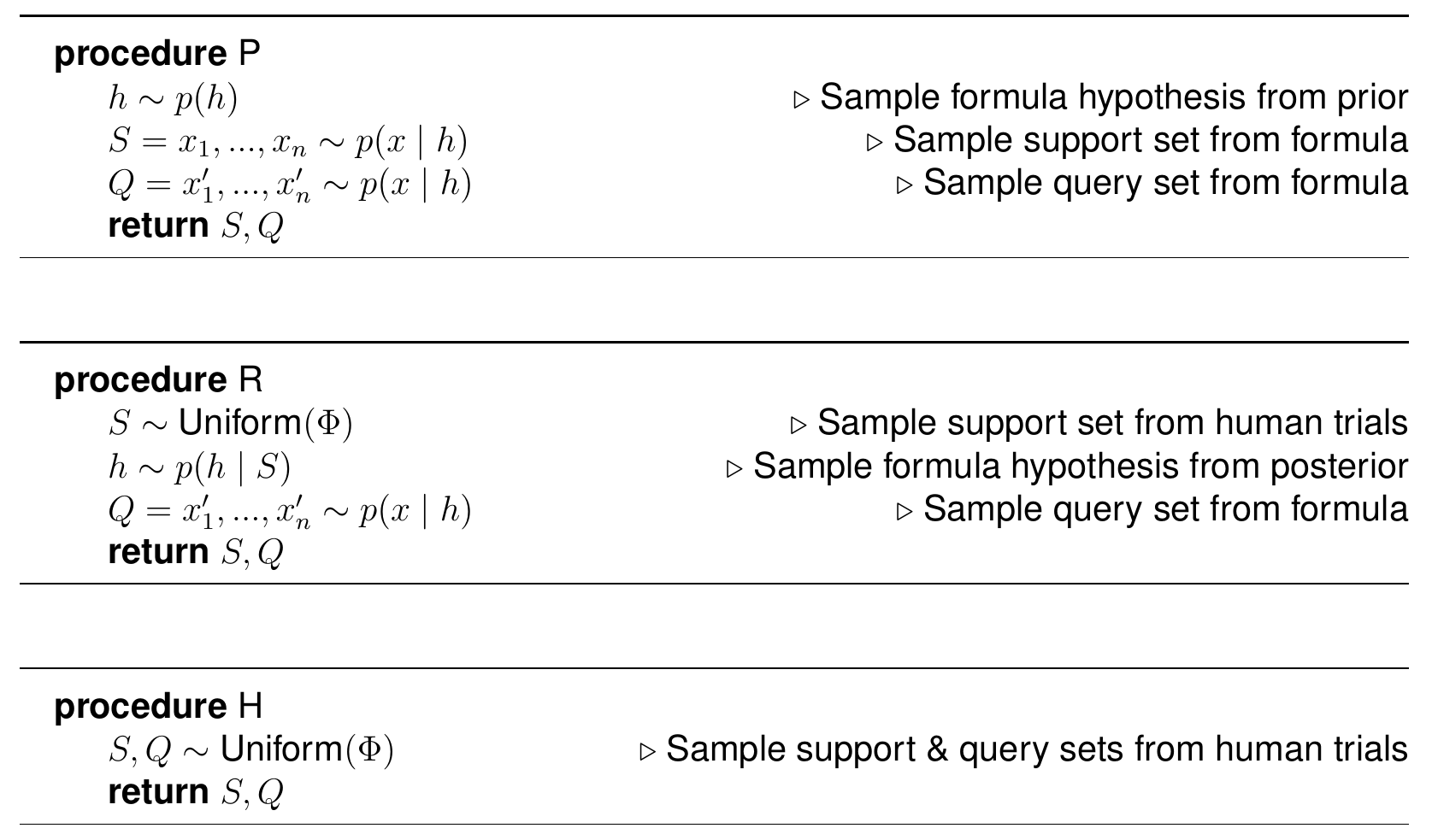}
    \caption{Data distributions for meta-learning.}
    \label{fig:gns_data_distributions}
\end{wrapfigure}
The objective of few-shot generation is to generate new tokens of a concept given a limited set of support exemplars. In the Bayesian setting, this task is modeled as sampling from the posterior predictive probability $p(y \mid X)$ of a new token $y$ given a support set $X = \{ x_1, ..., x_n \}$ consisting of $n$ exemplars.
Our GNS model provides a nonparametric analogue to the posterior predictive that can be estimated directly from training data, written $p(y \mid X) \approx f_{\theta}(y; X)$, where $f$ represents the model approximation parameterized by $\theta$.
To train GNS effectively, we borrow a paradigm from AI known as \textit{meta-learning} \shortcite{Hospedales2022}. Each input or ``episode" provided to the model consists of 1) a set of support tokens, a.k.a. exemplars, and 2) a set of query tokens for the model to evaluate.
Through these episodes the model \textit{learns-to-learn}, capturing overarching patterns that connect queries to support and learning to quickly grasp new concepts from exemplars.

\begin{figure}[ht!]
    \centering
    \includegraphics[width=0.95\hsize]{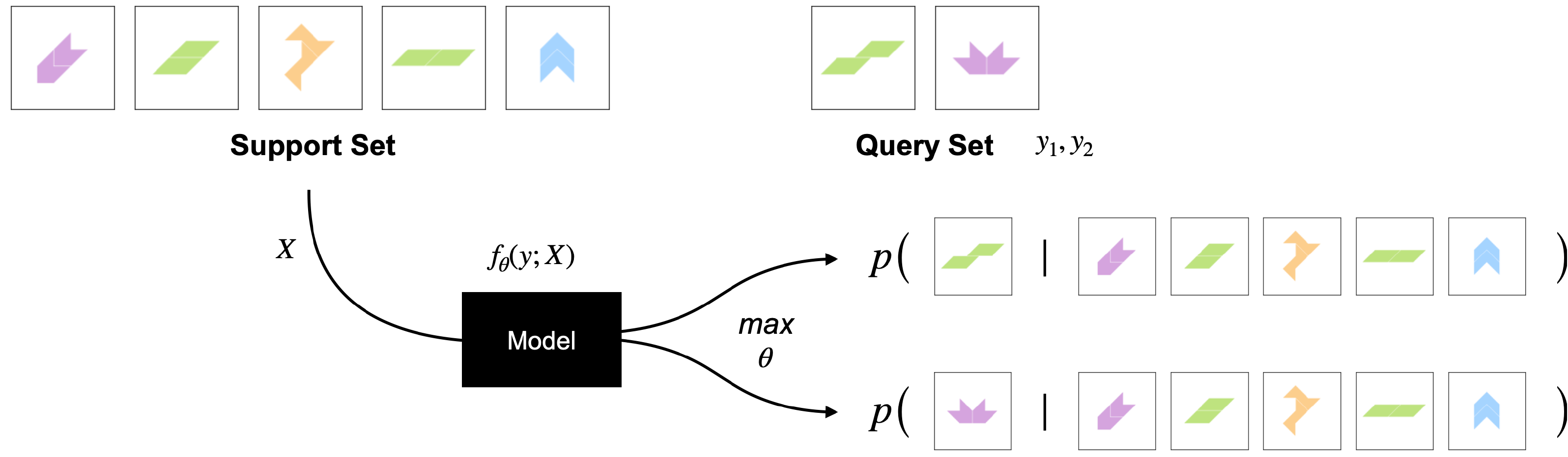}
    \caption{\textbf{Meta-learning episodes.} Each episode consists of 1) a support set $X$ of 1-6 examples that demonstrate the concept, and 2) a query set of additional tokens for evaluation $y_1,y_2,\dots$. The GNS model is trained to maximize the conditional log-likelihood of each query token given the support examples.}
    \label{fig:meta_learning_episodes}
\end{figure}

As with any statistical estimator that uses neural networks, our GNS model calls for a sizeable training dataset to avoid overfitting and ensure adequate generalization. The behavioral dataset from our generation task consists of just 155 trials in total, an insufficient amount of data by itself. To fill in the gap, we use our symbolic Bayesian model to bootstrap GNS training with a vast supply of synthetic meta-learning data. Specifically, we use the Bayesian model to form two distinct distributions for generating training data (Fig. \ref{fig:gns_data_distributions}). In the first distribution P, episodes are generated by first sampling a hypothesis $h$ from the prior and then sampling a support set $S$ and query set $Q$ from the likelihood $p(X \mid h)$ (Eq. \ref{ll}). In a second distribution R, episodes are generated by sampling a support set $S$ uniformly from the human experiment and then sampling query $Q$ via the posterior of the Bayesian model. In addition to these two synthetic data distributions, we also use real human data, dubbed distribution H, as part of the training mix. Episodes from H are sampled uniformly from the human experiment.

%In addition to the P, R and H distributions described above, we make use of one additional data distribution, C, which provides concerted training of two inductive biases that were prominent in the behavioral results from our generation task, and yet that the Bayesian program induction model was unable to capture.
%In trials where the support exemplars convey a partial pattern with one item apparently left out, participants exhibit two salient inductive biases. 
%First, people demonstrate a strong preference to \textit{complete the pattern} by generating the last unobserved item. We refer to this tendency as the ``completion bias,” and it occurs an aggregate 59\% of the time in applicable trials. When not completing the pattern, people exhibit a second, slightly weaker inductive bias that we denote the ``reconfigure bias.” This bias is characterized by a preference to take the familiarized primitives and reconfigure them into a novel, multi-part object. Participants exhibit the reconfigure bias an aggregate 14\% of the time in the studied trials. Our concerted bias training distribution, C, is designed to teach the GNS model these biases and help guide it to generalize in more human-like ways.
%Additional details about model training are provided in Appendix \ref{sec:appendix_training_gns_model}.
In addition to the P, R and H distributions described above, we make use of one additional data distribution, C, which provides concerted training of two inductive biases noted in Section \ref{sec_inductive_biases}(ii): the \textit{complete the pattern} bias and the \textit{reconfigure} bias. These two inductive biases are relevant in trials where the support exemplars convey a partial pattern with one item apparently left out (see examples in Fig.~\ref{fig:gns_inductive_biases_subset}). In all of the applicable trials, participants exhibit the completion bias an aggregate 59\% of the time, and they exhibit the reconfigure bias an aggregate 14\% of the time. Despite such high prevalence in the human data, these inductive biases are not well-explained by the Bayesian program induction model. Motivated by this shortcoming, we use the distribution C to provide direct training of these two inductive biases to the GNS model. Appendix \ref{sec:appendix_data_distribution_c} provides details about how episodes are generated from C.

% ------------------------------------------------------------------------

\subsection{Results}

% We set out to test whether the GNS model can successfully learn to generate new tokens from exemplars akin to the human behavioral task. In addition, we would like to study which of the proposed training distributions are most important to learning this task.
Our first simulation is designed to test whether the GNS model can successfully learn to generate new tokens from exemplars, and to determine what training distributions are most important for learning this task.
For the experiment, we constructed a test set of human data consisting of 1 randomly-selected trial from each trial type in the generation task (see Section \ref{sec:gen_exp} for details on trials \& trial types). The remaining 4 trials of each type are provided for model training. 
By reserving a portion of the human data for test time, we can use log-likelihood evaluations to assess whether the GNS model generalizes to novel trials with unseen behavioral data, and to compare the behavioral account of GNS to that of the Bayesian model.

\begin{table}[ht]
    \centering
    \begin{tabular}{lccc}
        \toprule
        Model & log-likelihood & t-statistic (p-value) \\
        \midrule
        %Bayesian      &  -4.859      &  - \\
        %GNS (P/R/H/C)  & \textbf{-4.479}   &  5.982 (0.000) \\
        %GNS (P/R/H)     &  -4.582     &  4.530 (0.000) \\
        %GNS (P/R)    &  -4.704       &  2.822 (0.005) \\
        %GNS (P)      &  -5.499      &  -4.485 (0.000) \\
        Bayesian      &  -4.741      &  - \\
        GNS (P/R/H/C)  & \textbf{-4.444}   &  6.197 (0.000) \\
        GNS (P/R/H)     &  -4.535     &  4.549 (0.000) \\
        GNS (P/R)    &  -4.645       &  2.490 (0.013) \\
        GNS (P)      &  -4.930      &  -2.739 (0.006) \\
        \bottomrule
    \end{tabular}
    \caption{Holdout log-likelihoods. For each model, the average log-likelihood per human token is reported in the first column. For each GNS model, we perform a paired t-test to test for improvement over the Bayesian model. The full GNS model, and all but one lesion model, show an improved behavioral fit over the Bayesian model, fortified by significant t-test results.}
    \label{tab:likelihood_results}
\end{table}

Our full GNS model, GNS (P/R/H/C), uses a mixture of all four training distributions described in the previous section. 
This represents our most comprehensive training environment, and we anticipate that the resulting model will outperform alternatives that receive only a subset of the proposed training distributions.
We test a series of these alternatives.
The first, GNS (P/R/H), receives all but the bias training distribution C.
In addition, we also evaluated two lesions that receive only synthetic data from the Bayesian model. 
One of these, GNS (P/R), receives data from both of the two synthetic generators. 
The other, GNS (P), uses only the forward-sampling modality P.
Each of our models is trained using minibatches of 60 meta-learning episodes (Appendix Section \ref{sec:appendix_training_gns_model}).

Log-likelihood results for held-out human data are shown in Table \ref{tab:likelihood_results}. 
When evaluating test log-likelihoods, we mix the model distribution with a naive lapse distribution using weight $\alpha$ that is independently tuned for each model (Section \ref{sec:appendix_gns_likelihoods}).
Our full GNS model, GNS (P/R/H/C), performs the strongest on held-out data and shows a considerable improvement in log-likelihood over the Bayesian model. The improvement is further validated by a significant paired t-test that looks at per-token difference in log-likelihood $\ell(\theta) - \ell(\theta_0)$ between the GNS model, $\theta$, and Bayesian model $\theta_0$ [t(336) = 6.197, p < 0.001].
After lesioning the bias training distribution, our GNS (P/R/H) model still exhibits a significant log-likelihood improvement over Bayesian program induction, although the gain is smaller.
The simplest lesioned model, GNS (P), performs the weakest on held-out data and does not show a significant improvement over the Bayesian model. This result matches our intuition: the space of possible episodes generated from P is vast, and so it is unlikely that the model will receive sufficient experience with the types of support sets that are relevant to our human experiment. Our second lesion, GNS (P/R), is the first to outperform the symbolic Bayesian model and show a statistically significant improvement in log-likelihood. Like GNS (P), this model is trained solely on synthetic data from the Bayesian model; however, the way that episodes are sampled in R---by selecting a support $S$ from the human experiment and then sampling query $Q$ from the Bayesian posterior---ensures that a sufficient amount of relevant training experience is provided.

\begin{figure}[ht!]
    \centering
    \includegraphics[width=0.75\hsize]{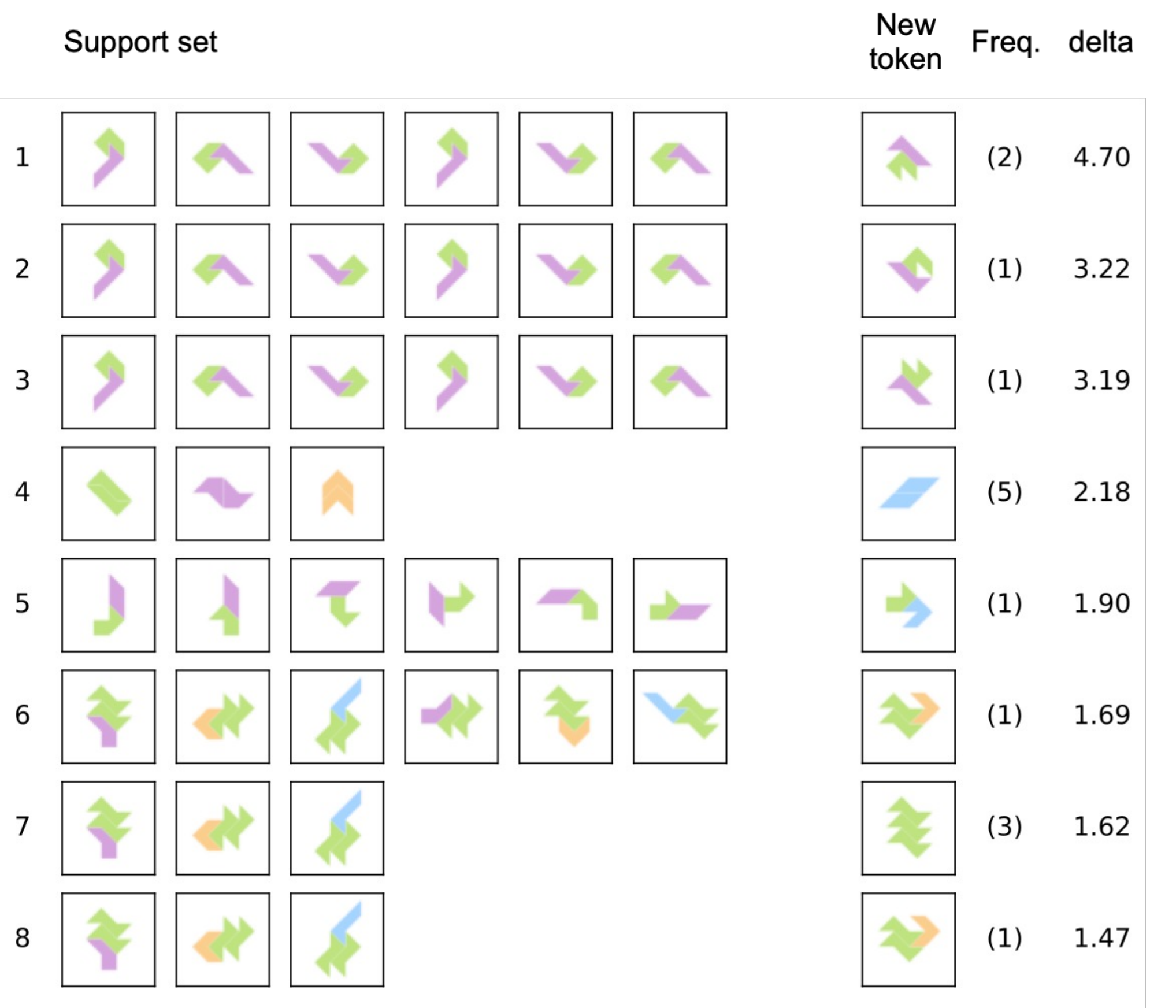}
    \caption{A subset of most-improved examples, measured by $\ell$(GNS) - $\ell$(Bayes).}
    \label{fig:gns_vs_bayes_subset}
\end{figure}

To help understand how and where our full GNS model outperforms Bayesian program induction, Fig. \ref{fig:gns_vs_bayes_subset} shows some of the top-performing examples where the log-likelihood improvement is largest (a more exhaustive set of best and worst examples is provided in Fig. \ref{fig:gns_vs_bayes}). 
The GNS model does particularly well with the two-part concept from rows 1, 2, and 3. In this trial, the size principle pushes the Bayesian model to assign most posterior weight to an attachment-specific hypothesis, so when a new token is shown with a different attachment, it loses out.
GNS also outperforms on the completion bias example from row 4, a result that is expected since the model receives explicit completion bias training from distribution C.
In row 5, the Bayesian model assigns a majority of posterior weight to a primitive-specific hypothesis, and it therefore suffers on the human-generated token that uses a new primitive.
The concept from rows 6, 7 and 8 has a salient visual compound that likely guides stronger generalization in human participants, but the Bayesian model is not aware of the compound. The GNS model, however, is capable of picking up on this visual pattern and mirroring human generalization.

\afterpage{\clearpage}
% \begin{figure}[ht!]
\begin{figure}[p]
    \centering
    \includegraphics[width=0.98\hsize]{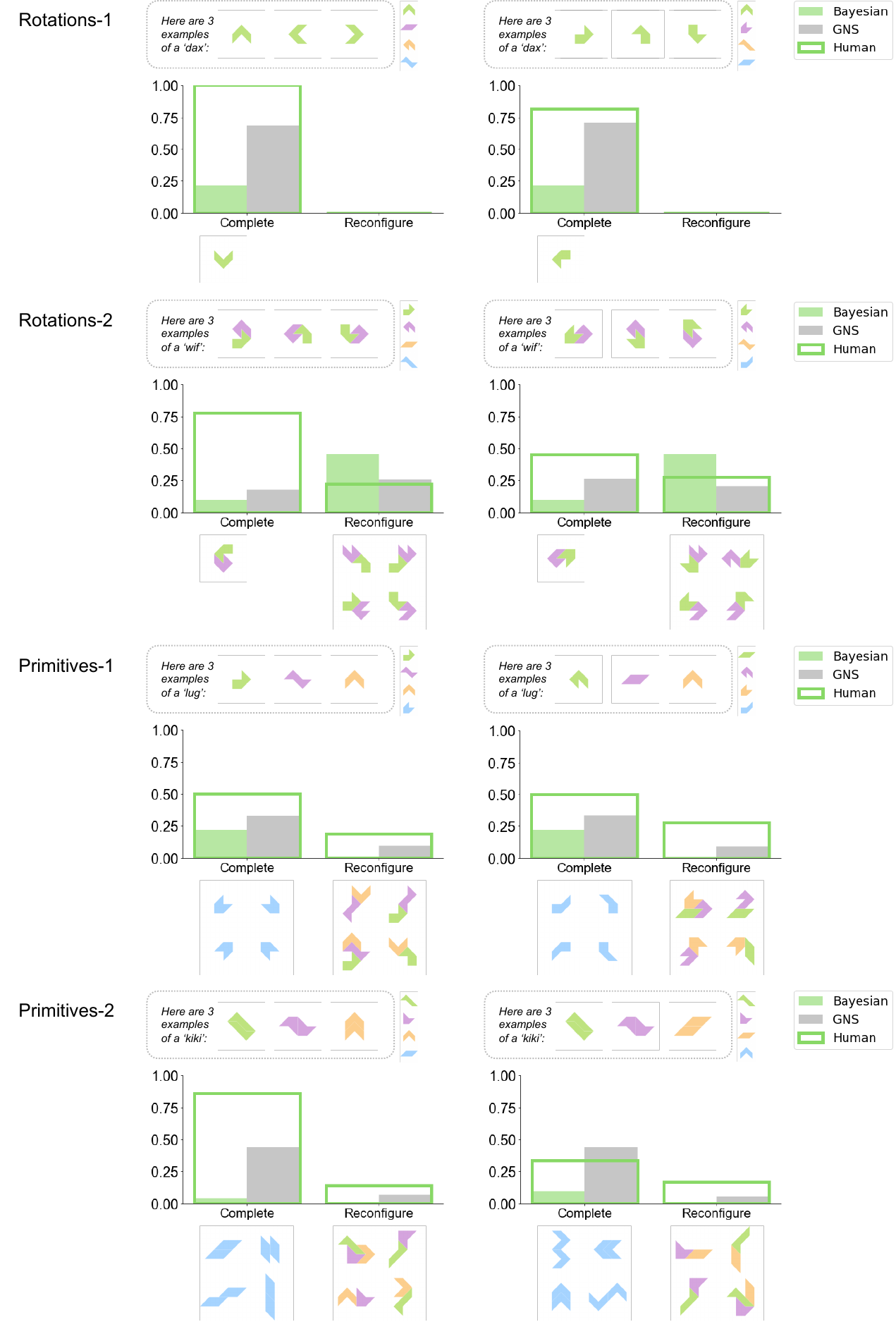}
    \caption{Inductive biases captured by the GNS and Bayesian models. Two trials are shown from each of four trial types with the partial-pattern property. Bars convey the marginal model probability of generating a new token that matches the target bias, and the empirical human frequency of doing so. In each trial, GNS exhibits a stronger completion bias vs. the Bayesian model that more closely matches human behavior. Moreover, the GNS model provides a closer match to human frequency for the \textit{reconfigure bias}, assigning a non-zero probability where Bayes does not and showing a more modest probability where Bayes overpredicts.}
    \label{fig:gns_inductive_biases_subset}
\end{figure}

To further understand how and whether the GNS model provides an improved account of human inductive biases, we conducted an additional simulation designed to give a more in-depth look at the \textit{complete-the-pattern} and \textit{reconfigure} biases discussed in Section \ref{sec_inductive_biases} \& \ref{sec:training_with_metalearning}.
We emphasize these two biases in particular because a) they are the most prevalent inductive biases that we find in the human behavioral data, and b) they are not currently well-explained by the Bayesian model.
To evaluate whether the GNS model can capture these biases, we created a test set with all 19 trials that contain the partial-pattern property discussed, as well as 7 other randomly-selected trials from the generation task. We then trained the full GNS (P/R/H/C) model using only the remaining trials for the human distribution H.
% Fig. \ref{fig:gns_inductive_biases} we provide a more detailed qualitative analysis of the GNS model by visualizing predictions from a collection of holdout trials.
Fig. \ref{fig:gns_inductive_biases_subset} shows the strength of the GNS model's inductive biases for a selection of test trials after training, comparing against both the Bayesian model and humans.
The Rotations-N trial type consists of N-part tokens with a rotation pattern, and Primitives-N consists of N-part tokens with a primitive assignment pattern.
People consistently exhibit a strong completion bias across different trial types, and the GNS model largely replicates this bias, showing a marginal probability for completion tokens that is often much closer to the human frequency compared with the Bayesian model.
In addition, the GNS model's reconfigure bias matches humans in strength more closely than the Bayesian model, showing more accurate probabilities where the Bayesian model overpredicts in Rotations-2 trials, and where it underpredicts in Primitives-1 trials.

\subsection{Experiment 3 discussion}

Experiment 3 demonstrates that generative neuro-symbolic (GNS) models can provide an effective means to understand and simulate human behavior in few-shot generation. When trained with a novel meta-learning scheme that mixes synthetic and real human data, our GNS model successfully mimics the symbolic Bayesian model and goes beyond to capture additional human biases that were not previously well explained. Our full GNS model shows a considerable improvement in likelihood of held-out participant data, and it provides an improved account for two salient inductive biases that participants exhibit: the ``complete-the-pattern" bias and the ``reconfigure" bias. In addition to these salient inductive biases, the GNS model also provides an account for a collection of one-off behaviors that do not fit into a larger bias category (Fig. \ref{fig:gns_vs_bayes_subset}).

\section{General Discussion}

Across a set of experiments, we study few-shot visual concept learning and generalization, with a focus on concepts that compose primitives together in a variety of different ways. We provide new empirical results---spanning both classification and generation tasks---on how people learn and generalize new visual concepts that better reflect the variety of ways parts combine in the real world. To model flexible visual composition, we develop a Bayesian program induction model that searches over different structured generative programs describing how parts can combine to best explain a set of exemplars (i.e., alien figures). The highest scoring programs can then be utilized for classifying or generating novel examples. This model provides a strong account of human categorization judgements, even in highly ambiguous trials, with a limited set of interpretable free parameters that offer insight into people's assumptions about invariance. For example, we find that people expect category membership to be invariant to rotations and changing part attachments, although these expectations can be updated in light of contradicting data. Additionally, the Bayesian program induction model can replicate these biases that human participants also demonstrate when generating novel alien figures, producing samples that are indicative of orientation invariance and attachment invariance (Fig.~\ref{biases}A\&B). 

Representing concepts as structured programs is one of the key principles for successful modeling in our compositional concept learning tasks. Structured programs provide the representational flexibility for modeling a rich variety of concepts, including those with tightly-constrained exemplars adhering closely to a specific part/relation pattern (Fig.~\ref{intro} first and third rows), or those with more widely varying exemplars with some defining characteristic (e.g., a part or set of parts), or those following abstract rules that require variable binding (Fig.~\ref{intro} last row) \shortcite{Marcus2003,overlan_learning_2017}. Human participants are not informed in advance what kinds of composition to expect, and thus it is essential for models to flexibly construct candidate programs (rather than assume a pre-defined set) in response to observations \shortcite{lake_human-level_2015}. The grammar used by our Bayesian program induction model is designed to produce programs that reflect a diverse set of visual compositions with its set of unique shape primitives and their rich space of relations. The Bayesian model also fluidly handles variable binding, producing concepts with various levels of abstractions. In contrast, the lesioned Bayesian no-Var model and the two alternative exemplar models applied to the categorization task fall short when reasoning with variables was required, and the Bayesian no-DP model struggles on trials where assumptions about defining parts are probed and tested.

We also observe more subtle human behavioral phenomena beyond the scope of the Bayesian program induction model's capabilities (Fig. \ref{fig:gns_inductive_biases}). Although incorporating more inductive biases into our existing Bayesian program induction model is possible through expanding the grammar to have more complex rules or through designing specialized likelihood functions that model correlations between tokens, such efforts involve potentially endless hand engineering to capture every nuance of human behavior. Instead, we demonstrate an alternative approach based on Generative Neuro-Symbolic (GNS) modeling that allows us to bootstrap the success of the Bayesian model and maintain flexible, explicit part composition while capturing additional behavioral nuances in a data-driven way. The resulting GNS model outperforms the Bayesian program induction model in terms of log-likelihood of generated human exemplars. GNS also helps to capture key behavioral phenomena missed by the Bayesian model, such as the ``complete-the-pattern'' bias that violates common likelihood assumptions used in Bayesian models of concept learning.

Although standard deep neural network models have difficulty with compositional generalization \cite{Fodor1988,Marcus2003,lake_building_2017,LakeBaroni2018}, we show that a hybrid approach, with a symbolic control flow that calls neural sub-routines to generate parts and their relations, can successfully model few-shot compositional learning and capture patterns of behavior missed by the Bayesian induction model. Although the current GNS model is an important step forward, it relies on the Bayesian model and other synthetic generators for training data, and as result, it may include some of the same biases and shortcomings of these parametric distributions. In future work, we would like to scale up the human experiment to provide enough training data directly probing the complete-the-pattern type inductive biases such that we can learn a GNS model more directly from human behavior. The current dataset includes only 155 trials in total, which is insufficient to train the neural network components considered in this article.

Despite the reported successes of our computational approaches, there are still aspects of human behavior in the current visual concept learning experiments not fully accounted for by our models. First, we observe a divergence of behavioral patterns between the categorization task and the generation task, which lead to distinct MAP values for a subset of the free grammar parameters fitted separately on the two tasks (Fig.~\ref{params}). Model performance suffers when parameter values were transferred from one task to another, leading to a decline in average per-token log-likelihood for Experiment 2. One possible driving forces of this divergence is the set of inductive biases unique to the generative task. For example, both complete-the-pattern biases are only found when participants are asked to generate their own novel alien figures (see Appendix \ref{sec:cat_vs_gen} for illustrations). There are multiple possible explanations for differences between classification and generation: one possibility is that generative tasks elicit richer behavior from participants that reveal additional assumptions; another possibility is that participants engage in additional reasoning about what makes a particularly ``good example'' of the concept rather than a random example, or what particular example the experimenter may be looking for. GNS, through its generative neural network components, could potentially provide the modeling power to capture these additional factors, although more work is needed to develop a complete theoretical account of the differences between categorization and generation behaviors.

We also found intriguing preliminary evidence that participants are sensitive to certain visual `motifs' that the Bayesian program induction model is blind to. In particular, people seem to be more visually attuned to compositions of shape primitives that are more symmetric and have smoother contours, and more easily perceived as gestalt entities than other randomly generated compositions (see an example in Appendix \ref{sec:motif}). Although not observed directly in our particular stimuli, certain motifs may also be particularly salient because of their connection with background knowledge \shortcite{Murphy2002}; for instance, observing exemplars with a shape that, by happenstance, resembles a fish's silhouette would likely influence participant judgements. The Bayesian program induction model is not well-equipped to account for these potential factors, as they are not evident from a concepts symbolic structure description. GNS, on the other hand, could in principle learn these factors via its neural network components, using either visual pre-training or large amounts of human behavior to acquire aspects of background knowledge. More computational work is needed to demonstrate this possibility and more empirical work is needed to understand the degree to which outside visual factors drive human generalization behavior.

%Importantly, the complete-the-pattern biases in Fig.~\ref{biases}C\&D identified in the current study are not uniquely present in the context of visual concept learning. \citeA{LakeLinzenBaroni2019} observed a similar effect when probing mutual exclusivity in compositional function learning: when there were only 2 possible token options for an unassigned name, one of which already associated to another name, people overwhelmingly preferred the leftover token to ``complete the pattern". This effect was weakened when the pool size of tokens increased. Further experiments can help elucidate the role of the complete-the-pattern type biases play in visual concept learning. 

There are many additional avenues for extending the computational modeling approaches pursued here. Extending models of compositional visual concept learning to naturalistic images and/or 3D models of real objects is one important direction, moving beyond synthetic stimuli studied here (alien figures; Fig.~\ref{intro}B right) to the types of everyday concepts people learn (bikes, vehicles, pairs of gloves, etc.; Fig.~\ref{intro}B left). We see the Bayesian program induction model as providing critical guidance regarding the key ingredients for moving forward---Bayesian inference over structured representations---while we see GNS as the most promising practical means of building models with these ingredients that also interface with noisier natural images (as neural networks are highly adept at doing) and capture complex correlational structure between parts that  fully symbolic models may miss. Extending models to capture how people learn compositions of functions, and how functions relate to object parts and visual appearances, is another critical extension. For instance, human understanding of the ``breakfast machine'' (Fig.~\ref{intro}A) is greatly enhanced by understanding how parts relate to functions (toaster, griddle, coffee maker, etc.) and how composing the parts relates to composing the functions. A complete account of human visual concept learning would thus need to relate form to function, either through inferring more sophisticated symbolic programs or through more data-driven, embodied learning that places objects in functional roles. We hope that the empirical and modeling findings presented here will inform future efforts for meeting these additional challenges.

\section{Acknowledgement}

The authors would like to thank Guy Davidson and A. Emin Orhan for helpful discussions and comments on an earlier version of this manuscript. We also thank members of the Human and Machine Learning Lab for thoughtful feedback during lab meetings at various stages of this project. 

This work was supported by NSF Award 1922658 NRT-HDR: FUTURE Foundations, Translation, and Responsibility for Data Science. Yanli Zhou was supported by the Meta AI Mentorship Program. Reuben Feinman was supported by a Google Fellowship in Computational Neuroscience.

\bibliographystyle{apacite}
\bibliography{references,library_clean} %-->reference list is on the template.bib file

\newpage
\appendix

\section{Full set of grammatical rules}
\begin{figure}[p]
\centering
\includegraphics[width=0.95\columnwidth]{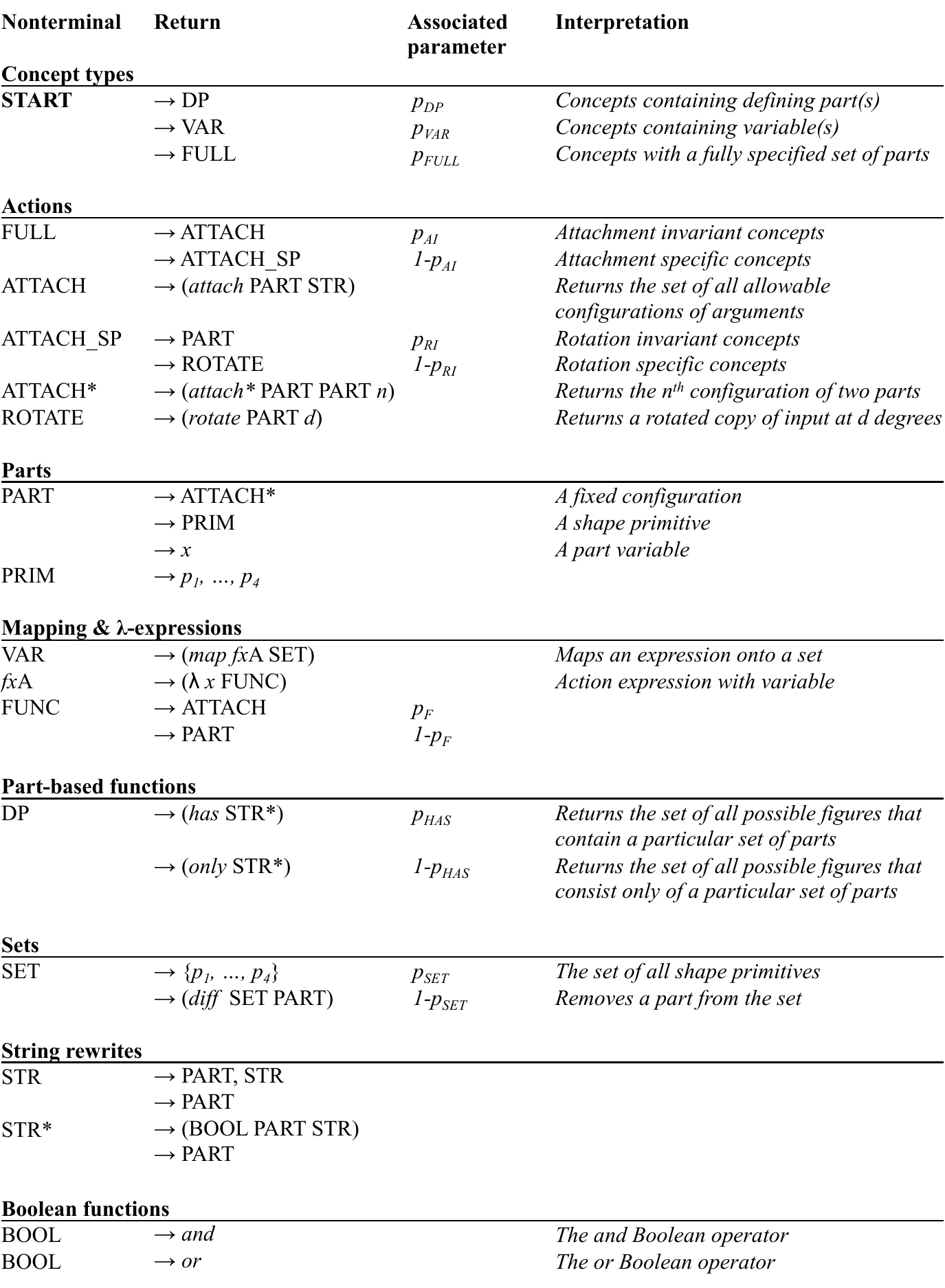}
\caption{{\bf Full set of grammatical rules.}
The full probabilistic context-free grammar used by the Bayesian program induction model to define the space of all possible alien figures. Non-terminals are indicated by uppercase letters. See text for details.}
\label{fullgrammar}
\end{figure}

Fig.\ref{fullgrammar} show the full set of expansion rules associated with the grammar used by the Bayesian program induction model. From the START symbol, nonterminal nodes are expanded into their downstream nonterminal until a terminal node has been reached. The set of free grammar parameters fitted to the human datasets are also indicated.

\afterpage{\clearpage}
% \begin{figure}[ht!]

\section{Full set of trial types and categorization results} 
\begin{figure}[!ht]
\centering
\includegraphics[width=1\columnwidth]{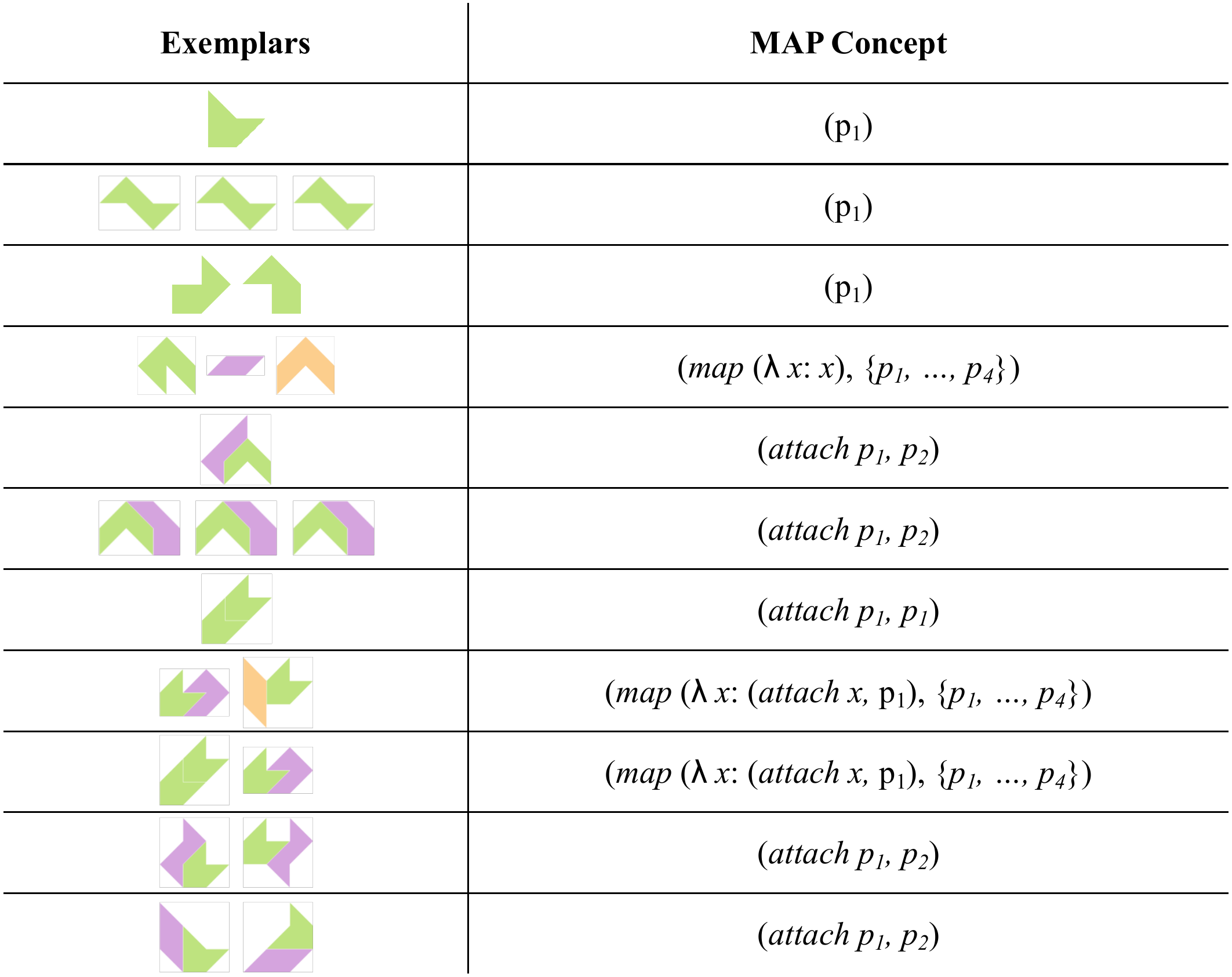}
\caption{{\bf Experiment 1a trials and their associated MAP concepts.}
Left column shows the examples of exemplar sets shown to the participants on each trial of Experiment 1a. Note that the set of shape primitives were randomized per participant per trial. Right column shows the hypotheses assigned the highest posterior probabilities by the Bayesian program induction model in each trial of Experiment 1.}
\label{exp1}
\end{figure}

\begin{figure}[!ht]
\centering
\includegraphics[width=1\columnwidth]{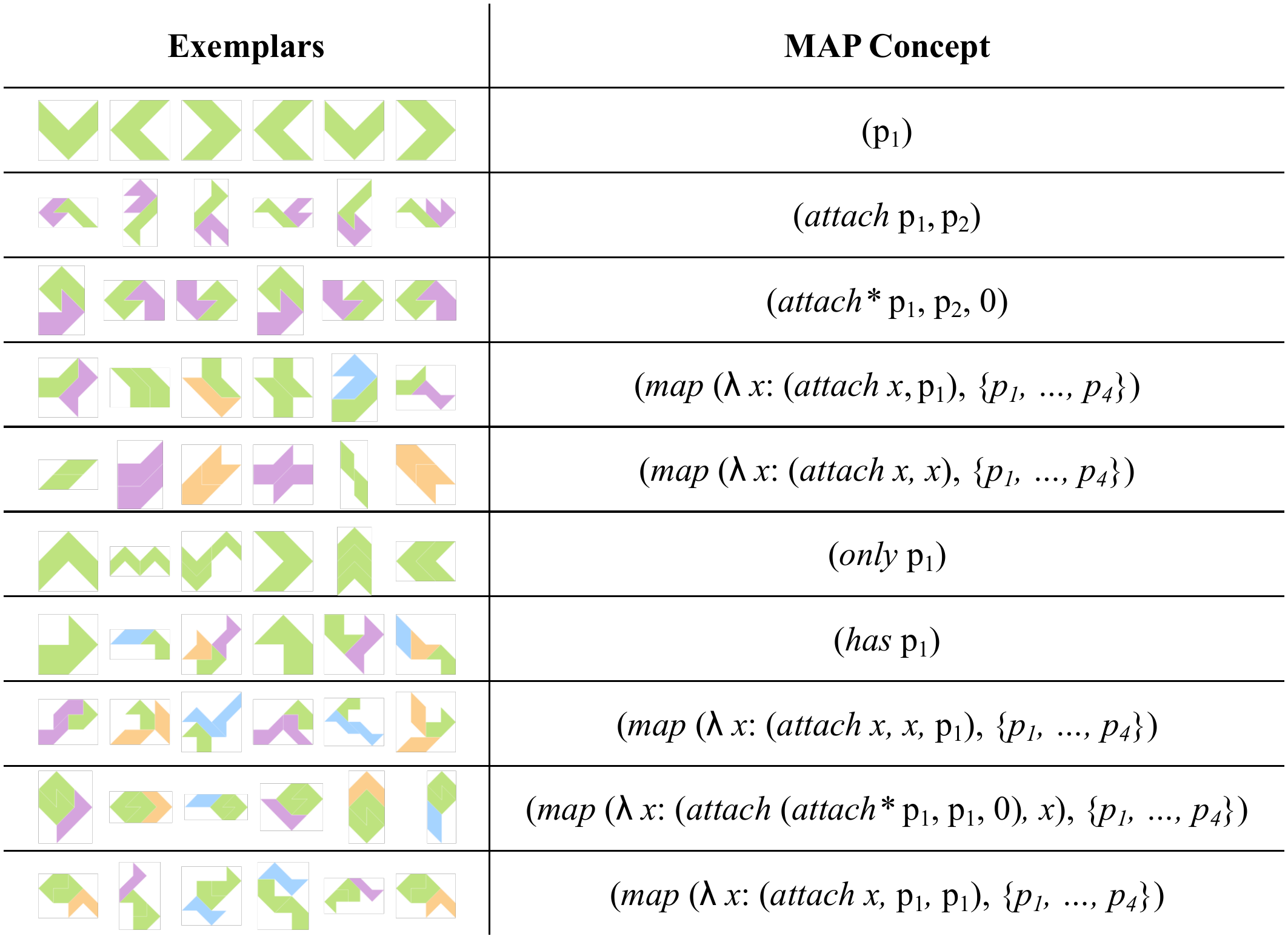}
\caption{{\bf Experiment 1b trials and their associated MAP concepts.}
Left column shows the examples of exemplar sets shown to the participants on each trial of Experiment 2. Participants assigned to the 3-exemplar condition were shown the three left most exemplars for every trial type, and participants in the 6-exemplar condition were shown all six exemplars. Right column shows the hypotheses assigned the highest posterior probabilities by the Bayesian program induction model in each trial of Experiment 1b.}
\label{exp2}
\end{figure}

\begin{figure}[p]
\begin{center} 
  \includegraphics[width=0.99\hsize]{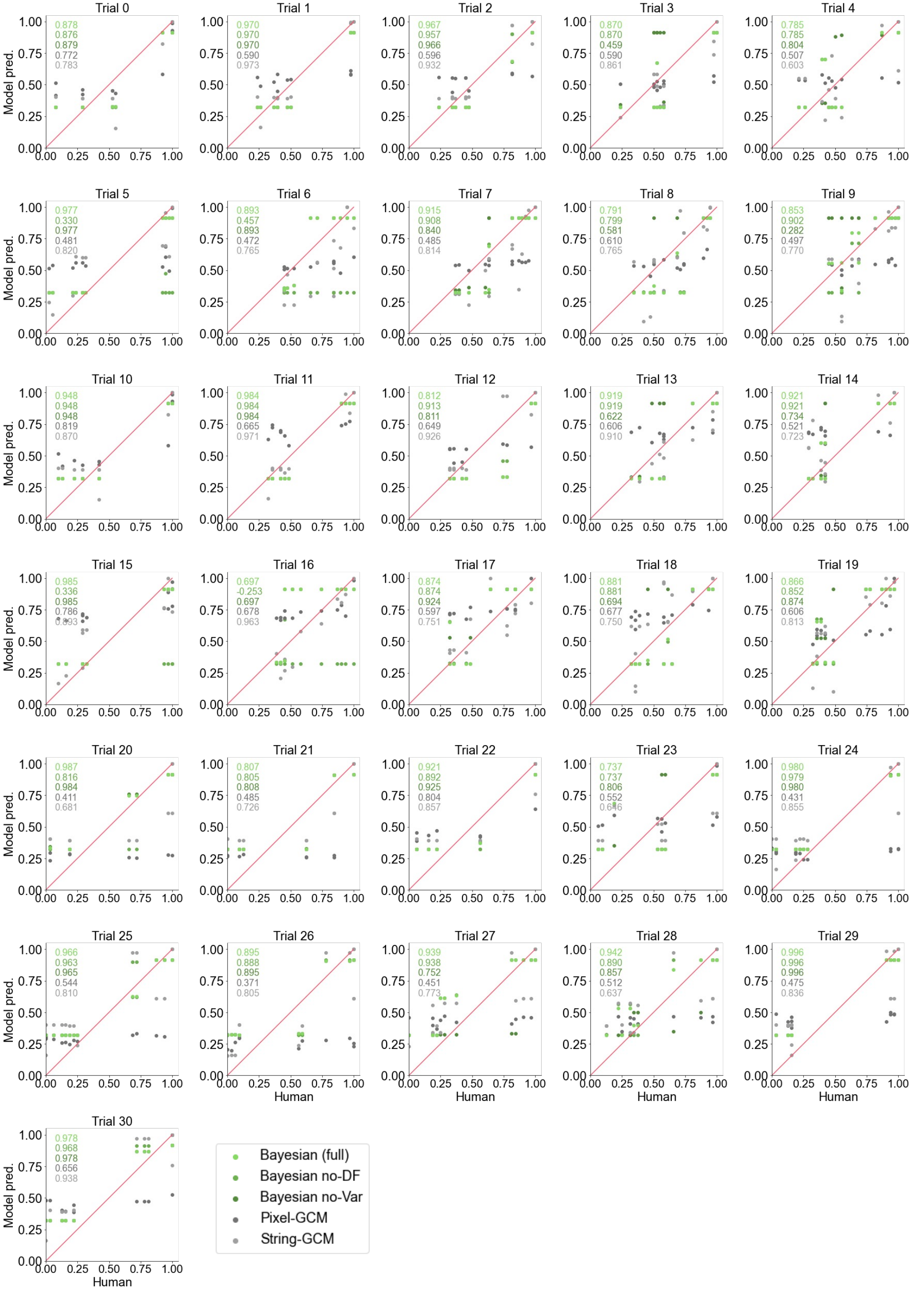}
  \caption{\textbf{Full set of categorization results.}  Comparison between human responses and model predictions for each trial type of the categorization experiment. The set of exemplars participants observed for each trial is shown above each scatter plot. Each dot in a scatter plot indicates the probability of responding `Yes' for each test item. Human-model correlations are also shown for each trial and each model.}
  
\label{scatters}
\end{center} 
\end{figure}

Fig.\ref{exp1} and Fig.\ref{exp2} show the full set of trial types participants were tested on in Experiment 1a and 1b, along with the maximum-a-posteriori (MAP) concept associated with each trial. Trial types are shown with one possible assignment of shape primitives, and participants saw other possible random assignments of shape primitives. Fig.\ref{scatters} shows the full set of Experiment 1 results summarizing correlations between human judgments and model predictions per trial type and model. 

The two sets of trial types were combined and shown to participants in the generation task (Experiment 2). In the generation task, participants were also assigned to either the 3-exemplar condition or the 6-exemplar condition for all Experiment 1b trial types.

\afterpage{\clearpage}
% \begin{figure}[ht!]

\section{Bayesian model parameter fitting}\label{fitting_procedure}

Given the behavioral data collected in our experiments, we are interested in finding the set of grammar parameters that most likely generated participants' response patterns. Formally, we would like to infer the probability of the set of parameters of interest, given human response data in Experiment 1: $\argmax_{\vec{\theta},\alpha, \beta}  P(L | X; \vec{\theta}, \alpha, \beta)$, where $\vec{\theta}$, $\alpha$ and $\beta$ are parameters of the learning model and $L$ is the set of human assigned labels to test items; and given human generations in Experiment 2: $\argmax_{\vec{\theta}, \alpha}  P(Y | X; \vec{\theta}, \alpha)$, where $Y$ is the set of human generated tokens.  We only considered grammar parameters that are psychologically meaningful (e.g. parameters that encode participants' preferences for \emph{orientation invariance} and \emph{configuration invariance}),  and we fixed the rest of expansions to have uniform probabilities. We discuss the implications of the fitted values of these parameters in the Results section. 

In addition to the set of grammar parameters and the likelihood parameters $\alpha, \beta$, we also included two more free parameters in $\vec{\theta}$, a prior temperature $T_p$, and a likelihood temperature $T_l$ which control the strength of the prior in Eq.~\ref{prior} and likelihood in Eq.~\ref{ll} by raising them to the $1/T$th power, respectively.  By implementing the prior temperature parameter we control the overall confidence of the prior model: the prior becomes increasingly uniform as $T_p$ approaches higher values, assigning less preferential probabilities to shorter programs and vice versa. The likelihood temperature parameter adjusts the strength the of size principle effect: with lower values of $T_l$, the likelihood becomes more sensitive to the size of hypotheses and this sensitivity weakens as $T_l$ increases. 

Based on the approximate hypothesis space $\hat{\mathcal{H}}$, we re-normalized the temperature-adjusted prior distribution to be $\hat{P}(h;\vec{\beta}, T_p) \propto P(h; \vec{\beta})^{1/T_p} = \frac{1}{Z(\vec{\beta}, T_p)} P(h;\vec{\beta})^{1/T_p}$, we subsequently re-normalized the posterior distribution after likelihood temperature adjustment. The posterior distribution $\hat{P}(h\in \hat{\mathcal{H}} |X)$ becomes:

\begin{equation}\label{posterior}
\begin{split}
\hat{P}(h\in \hat{\mathcal{H}} |X) & = \frac{\frac{1}{Z(\vec{\beta}, T_p)} P(h|\vec{\beta})^{1/T_p} P(X|h)^{1/T_l}}{\sum_{h'\in \hat{\mathcal{H}}} \frac{1}{Z(\vec{\beta}, T_p)} P(h'|\vec{\beta})^{1/T_p} P(X|h')^{1/T_l}}\\
& = \frac
{\frac{1}{Z(\vec{\beta}, T_p) } P(h|\vec{\beta})^{1/T_p} P(X|h)^{1/T_l}}
{\frac{1}{Z(\vec{\beta}, T_p) } \sum_{h'}  P(h'|\vec{\beta})^{1/T_p} P(X|h')^{1/T_l}}\\
&= \frac
{ P(h|\vec{\beta})^{1/T_p} P(X|h)^{1/T_l}}
{\sum_{h'}  P(h'|\vec{\beta})^{1/T_p} P(X|h')^{1/T_l}}
\end{split}
\end{equation}

Together, the optimization problem for categorization judgments in Experiment 1 becomes:

$$\argmax_{\bm{\vec{\theta}}}  P(L | X; \bm{\vec{\theta}}) = \argmax_{\bm{\vec{\theta}}}  \prod_t^T \prod_j^m \binom{n}{k}p_j^k(1-p_j)^{n-k},$$

where $p_j = P(l_{y,j} | X_t)$, the probability that the label $l_{y,j}$ of the $j$th test item $y_j$ is consistent with the set of exemplars $X_t$ on trial $t$, as calculated in Eq.~\ref{response1}; $n$ is the number of participant responses collected for each trial while $k$ is the number of responses such that $l_{y,j} = 1$; and $\bm{\vec{\theta}} = \{\vec{\theta}, \alpha, \beta, T_p, T_l\}$. 

And for the generation data in Experiment 2:

$$\argmax_{\bm{\vec{\theta}}}  P(Y | X; \bm{\vec{\theta}}) = \argmax_{\bm{\vec{\theta}}}  \prod_t^T \prod_i^n P(y_{i} | X_t; \bm{\vec{\theta}}),$$ 

where $y_{i}$ is the $i$th participant generated token given observation $X_t$ for trial $t$, and $\bm{\vec{\theta}} = \{\vec{\theta}, \alpha, T_p, T_l\}$.

\begin{figure}[ht!]
\begin{center} 
  \includegraphics[width=0.75\linewidth]{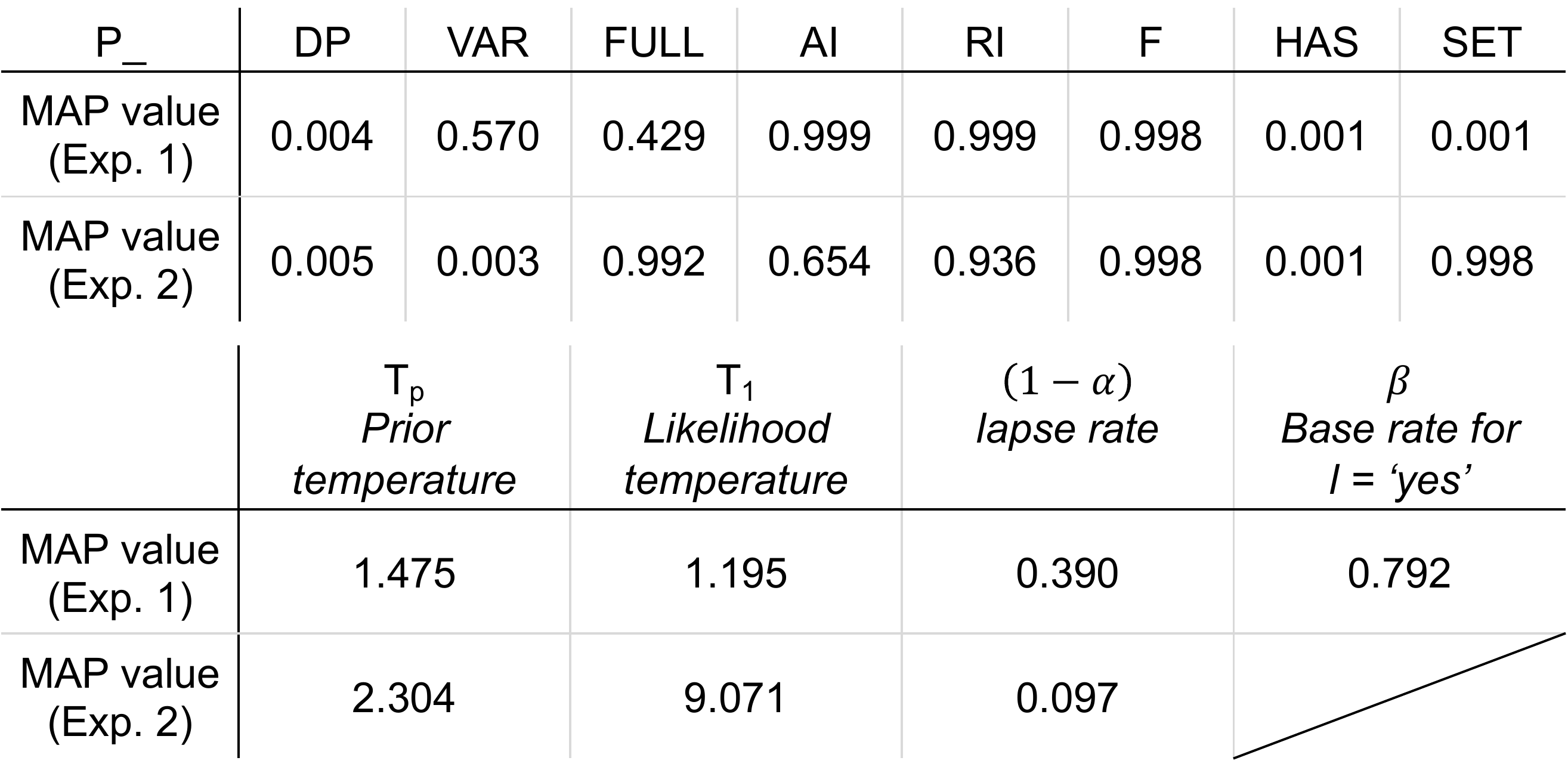}
  \caption{\textbf{Fitted parameters values of the Bayesian program induction model.} Top row shows the MAP parameter values when the model was fitted to categorization task (Experiment 1) data; the bottom row shows parameter fits for the generation task (Experiment 2).}
\label{params}
\end{center} 
\end{figure}

Free parameters are fitted via a sequential least squares programming algorithm (SLSQP); MAP parameter values for Experiment 1\&2 are reported in Fig.~\ref{params}.

% re-normalized the temperature-adjusted prior distribution to be $\hat{P}(h \in \hat{\mathcal{H}};\vec{\theta}, T_p) \propto P(h\in \hat{\mathcal{H}}; \vec{\theta})^{1/T_p}$ for each trial $t$, and we subsequently re-normalized the posterior distribution after likelihood temperature adjustment. The posterior distribution $\hat{P}(h\in \hat{\mathcal{H}} |X_t)$ becomes:

% \begin{equation}\label{posterior}
% \begin{split}
% \hat{P}(h\in \hat{\mathcal{H}} |X_t ) &\propto P(h\in \hat{\mathcal{H}})P(X_t|h\in \hat{\mathcal{H}}) \\
% &\propto \hat{P}(h;\vec{\theta}, T_p) P(X_t|h\in \hat{\mathcal{H}})^{1/T_l}
% \end{split}
% \end{equation}

\section{Null token distribution }\label{p0}

We model the null distribution of the tokens $P^{0}(x)$ while taking into account the complexity of tokens. Intuitively, more complex tokens should have lower probabilities: as we increase the complexity of a token by including more parts and attachments, the number of possible configurations increases exponentially, and thus the probability for any particular configuration should be smaller than that of a simpler token. The pseudo code below illustrates how a token $x$ is sampled and how its associated probability $P^{0}(x)$ is calculated. 

\begin{algorithm}
\caption{Generate a token $y$ from the null distribution $P(y)$. The cardinality of each uniform distribution depends on previous variates; for example, the number of valid relations $r_2$---i.e. the number of ways part 2 can attach to existing objects---depends on the primitives sampled for $p_1$ and $p_2$.}
\begin{algorithmic}
\item[]
\Procedure{GenerateToken}{}
    \State $p_1 \sim \text{Uniform}$ \Comment{Sample primitive for first part}
    \State $r_1 \leftarrow null$ \Comment{Null first relation}
    
    \For{$i = 2...T_{max}$}
        \State $p_i \sim \text{Uniform}$ \Comment{Sample primitive for $i^{th}$ part}
        \If{$p_i = terminate$}  \Comment{Check termination}
            \State break
        \EndIf
        \State $r_i \sim \text{Uniform}$ \Comment{Sample relation for $i^{th}$ part}
    \EndFor
    \State \textbf{return} $\{ p_0, r_0, ..., p_T, r_T \}$  
\EndProcedure
\end{algorithmic}
\end{algorithm}

% \section{Examples of inductive biases}
\section{Additional behavioral results}

\subsection{Divergence of behavior in Experiment 1\&2}\label{sec:cat_vs_gen}

\begin{wrapfigure}{r}{0.5\textwidth}
 \centering
  \includegraphics[width=\linewidth]{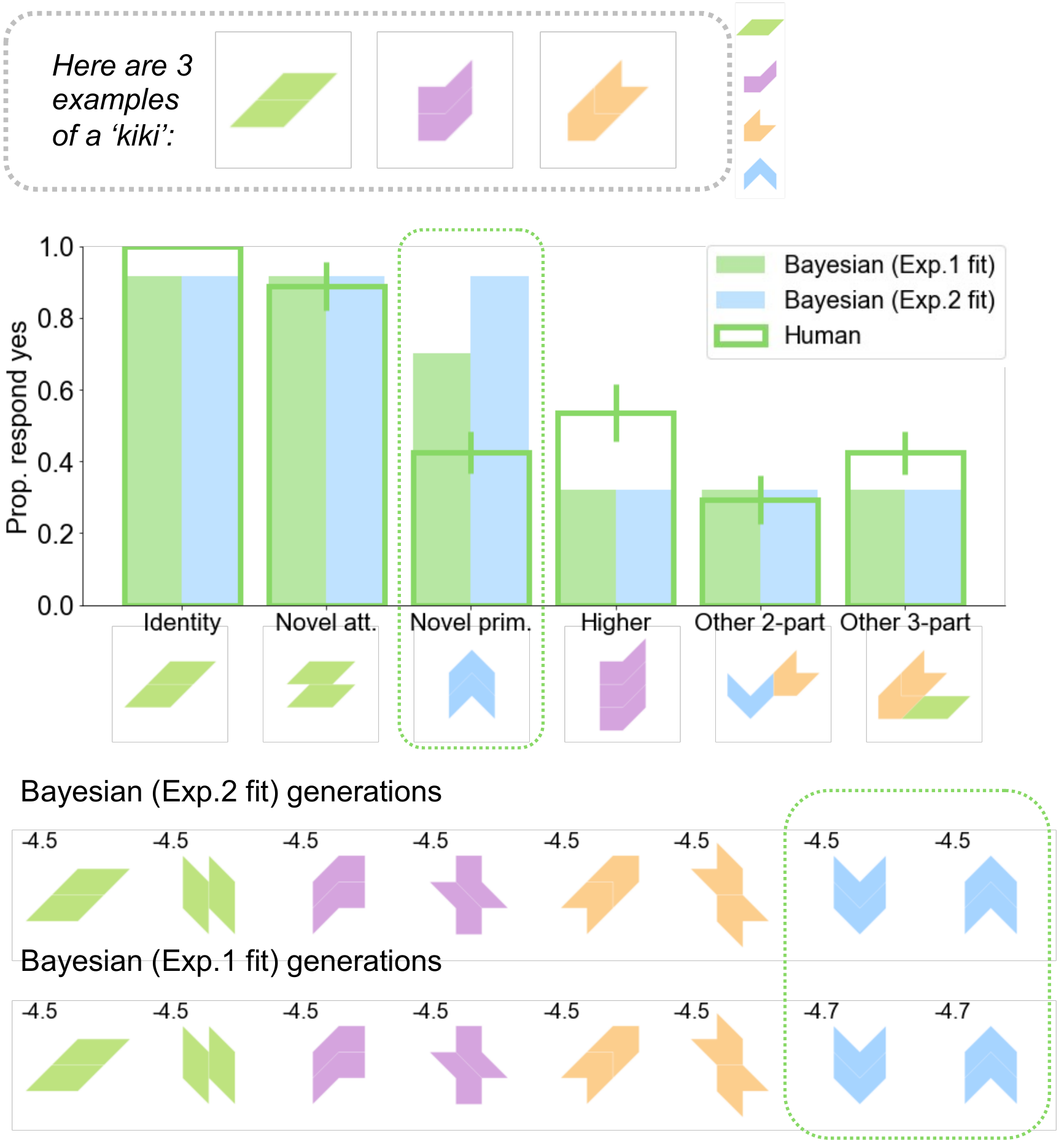}
  \caption{\textbf{Example trial suggesting divergence of behavior across tasks.} Bar plot shows model predictions for different test item types when grammar parameters are either fitted directly on categorization data (Exp.1 fit) or transferred from the generation task (Exp.2 fit). Bottom rows show model generations with their predicted (log) probabilities when grammar parameters are either fitted directly on generation data (Exp.2 fit) or transferred from the categorization task (Exp.1 fit).}
\label{fig:transfer}
\end{wrapfigure}

Human response patterns show qualitative divergence on a number of trials between the categorization and generation, leading to distinct MAP values for a subset of the grammar parameters in Fig.~\ref{params} across tasks, and suggestive of additional assumptions participants bear when asked to generate their own alien figures. For example, \textit{complete-the-pattern} biases are uniquely identified in the generation task, while interestingly, the categorization results reports an opposite effect. That is, when a similar all-but-one pattern was tested in categorization experiments, we see a slight drop in generalizations to test items that would ``complete the pattern" in comparison to the test items that would not complete a pattern but have been observed as a whole or a part in the exemplar set (see Fig.~\ref{fig:transfer}, highlighted bar). When plugging in the MAP values fitted for the generation task, the model assigns equally high probabilities for conceptually consistent test items with a novel primitive, whereas both participants and the Bayesian model fitted for the categorization data show a decline in generalization. Conversely, when asked to generate new examples based on observations, the Bayesian model with transferred parameters produces tokens with the primitive that completes the pattern at a lower probability (see Fig.~\ref{fig:transfer}, highlighted samples), a behavior in direct opposition with what we observed in human generation data.

\subsection{Sensitivity to ``visual motifs"}\label{sec:motif}

\begin{figure}[ht!]
\begin{center} 
  \includegraphics[width=0.75\linewidth]{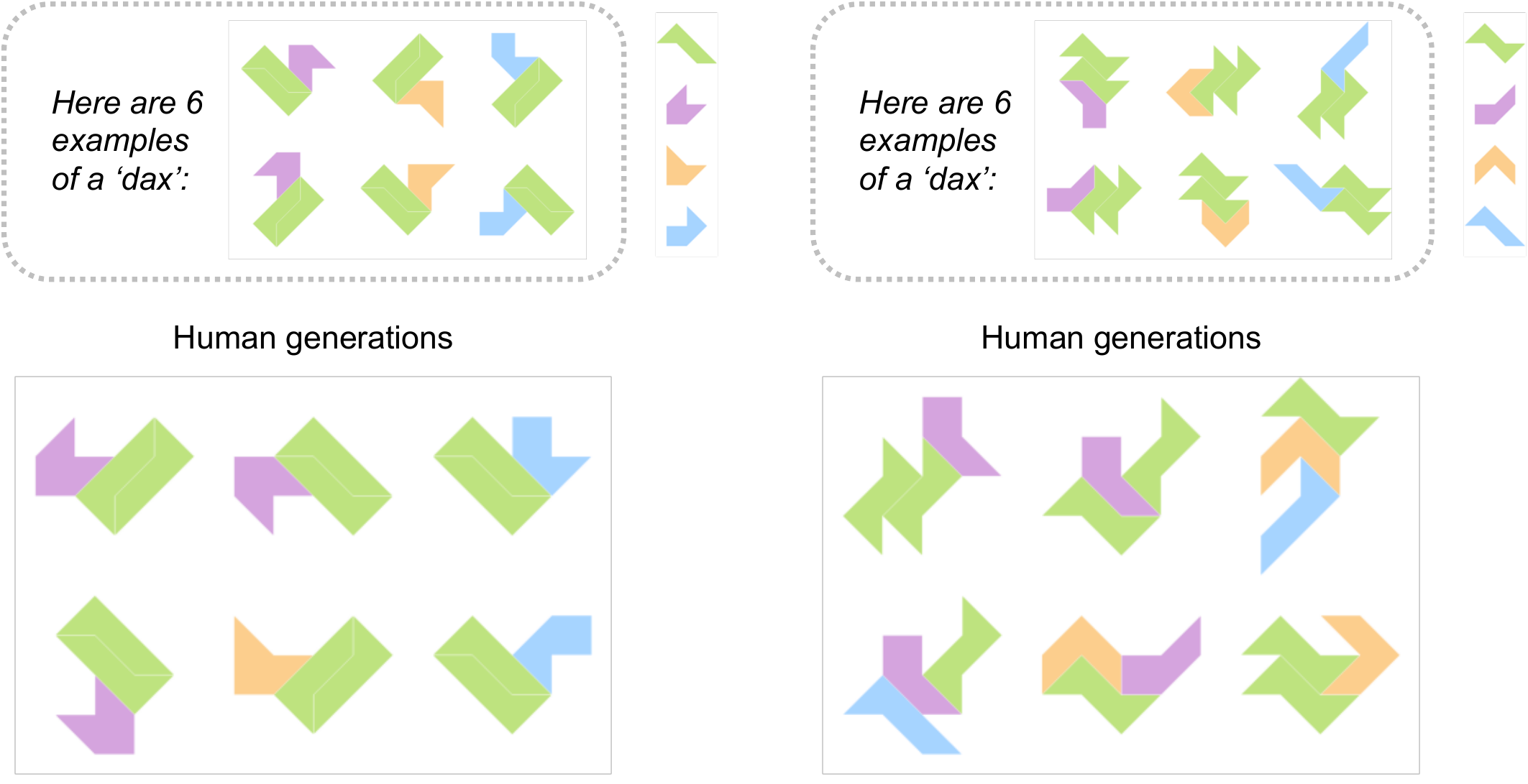}
  \caption{\textbf{Example of human sensitivity to visual motifs.} Two different random primitive assignments are shown for the same trial type. The green common subpart on the left happens to adopt a familiar rectangular form, while the green common subpart on the right has a more irregular outline. Generated tokens suggest that humans are more visually attuned to the more salient subpart on the left.}
\label{fig:motif}
\end{center} 
\end{figure}

We find evidence for sensitivity to primitive compositions that are more visually salient, which are usually highly symmetrical with familiar forms. For example, when the set of exemplars show a common subpart with a familiar, easily identifiable form, participants are more likely to generate tokens consistent with the underlying concept (Fig.~\ref{fig:motif}).

% \begin{wrapfigure}{r}{0.7\textwidth}
%     \centering
%     \includegraphics[width=0.69\textwidth]{images/top5acc.pdf}
%     \caption{Top-5 generation accuracy. \todo{moving to supplementary materials} }
%     \label{fig:top5acc}
% \end{wrapfigure}

% \input{appendix_gns.tex}

\section{GNS Model}

% ------------------------------------------------------------------------------------------------

\begin{figure}[ht!]
    \centering
    \includegraphics[width=1.00\hsize]{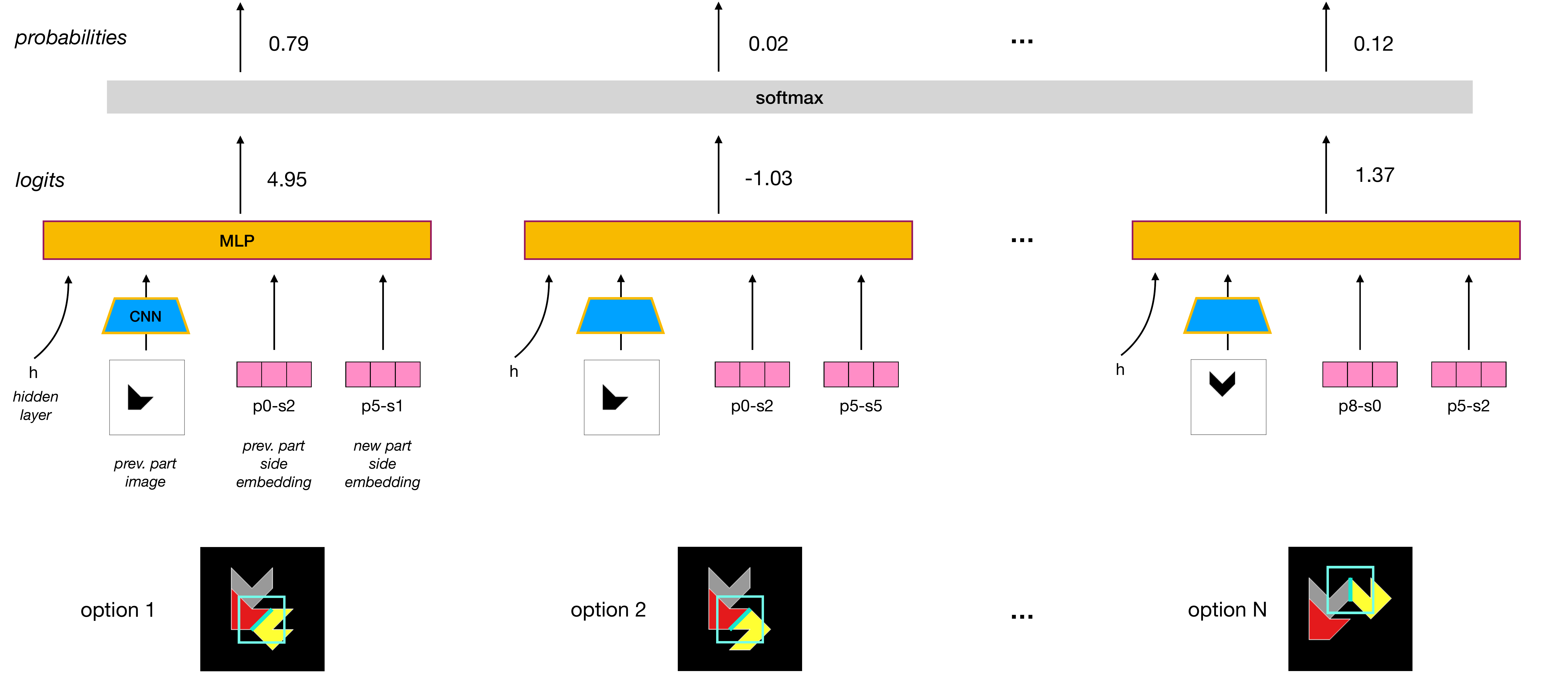}
    \caption{Relation prediction architecture used in GNS subroutine \texttt{GenerateRelation}.}
    \label{fig:gns_relation_prediction_architecture}
\end{figure}

\subsection{Relation architecture}
\label{sec:appendix_gns_relation_architecture}

The GNS model uses polygon attachments as a model of relations between parts in an alien figure. Each relation $r_i = \{j, s_j, s_i\}$ encompasses 3 unique choices which together specify an attachment. The first is the choice of attachment part, represented by index $j$, selected from the set of all previous parts. Second and third are the choice of polygon side for the attachment part $s_j$, and for the current part $s_i$. These choices convey which polygon sides will be touching when the two polygons are connected to one another.

To predict the next relation $r_i$, GNS uses a neural network as an energy function to score every combination of values $\{j, s_j, s_i\}$ (Fig. \ref{fig:gns_relation_prediction_architecture}). The choice of attachment part $j$ is conveyed by a binary image of the isolated part, which is processed to a hidden embedding by a CNN. The side choices $s_j$ and $s_i$ are each conveyed by a discrete embedding from a learnable dictionary with one entry for every side of every primitive polygon, indexed as $e[c_i, s_i]$. Each of these inputs is concatenated and fed to the neural network, which returns a scalar energy that represents the unnormalized log-probability of choosing this combination.

% ------------------------------------------------------------------------------------------------
\subsection{Training the GNS model}
\label{sec:appendix_training_gns_model}

Our full GNS model and all lesions are training using minibatches of 60 meta-learning episodes. The composition of data distributions for each lesion is provided in Table \ref{tab:minibatch_compositions}. 
%For our full model, GNS (P/R/H/C), the C distribution further decomposes into... 
The number of support examples in an episode is sampled uniformly between 1-6 at each iteration, and the number of query examples is fixed at 5.
Models are trained to maximize the log-likelihood (minimize log-loss) of the query examples conditioned on support. Training proceeds for 40,000 batch iterations using the Adam optimizer with cosine learning rate annealing.
For each GNS model, we train 4 different models with different random initialization. In subsequent evaluations, we use the average log-likelihood from all 4 seeds as the overall log-likelihood.

\begin{table}[ht]
    \centering
    \begin{tabular}{lccc}
        \toprule
        Model & Composition \\
        \midrule
        GNS (P)         &  60 \\
        GNS (P/R)       &  40/20 \\
        GNS (P/R/H)     &  30/15/15 \\
        GNS (P/R/H/C)   &  20/10/10/20 \\       
        \bottomrule
    \end{tabular}
    \caption{Minibatch compositions for GNS model training.}
    \label{tab:minibatch_compositions}
\end{table}

\subsubsection{Data distribution C}
\label{sec:appendix_data_distribution_c}

The C distribution is designed to help teach the complete-the-pattern and reconfigure biases, two inductive biases that are relevant in trials with the partial-pattern property. To generate episodes from C, we first sample a trial type from the four partial-pattern types: Rotations-1, Rotations-2, Primitives-1, Primitives-2. Next we sample a support set $S$ by selecting 3 tokens from the trial type that make a partial-pattern. Finally, to construct the query set $Q$ we sample completion items with probability $p_a=0.59$, reconfigure items with probability $p_b=0.14$, and alternate ``noise" tokens with the remaining probability mass. The values of $p_a$ and $p_b$ are set to mirror the empirical human frequencies for each bias.

% ------------------------------------------------------------------------------------------------
\subsection{Likelihood analysis}
\label{sec:appendix_gns_likelihoods}

All token likelihoods that we report for the GNS model are marginal image likelihoods. By default, the GNS model computes the likelihood of a \textit{latent program} or a \textit{token string}, i.e. a sequence of parts and relations $\{c_1, r_1, ..., c_N, r_N\}$. There is a many-to-one mapping from these latent programs to images; to obtain the marginal likelihood of a token image, we sum the individual likelihoods from all programs that yield the target image.

For both the GNS model and the Bayesian model, we fit a lapse parameter $\alpha$ that mixes the model distribution $p(y \mid X)$ with a null distribution $q(y)$ to produce a final distribution $\tilde{p}(y \mid X) = (1 - \alpha) \cdot p(y \mid X) + \alpha \cdot q(y)$.
We use the complexity-based null distribution $q(y) = P^{0}(y)$ discussed in Appendix \ref{p0}.

% ------------------------------------------------------------------------------------------------

\begin{figure}[ht!]
    \centering
    \includegraphics[width=0.97\hsize]{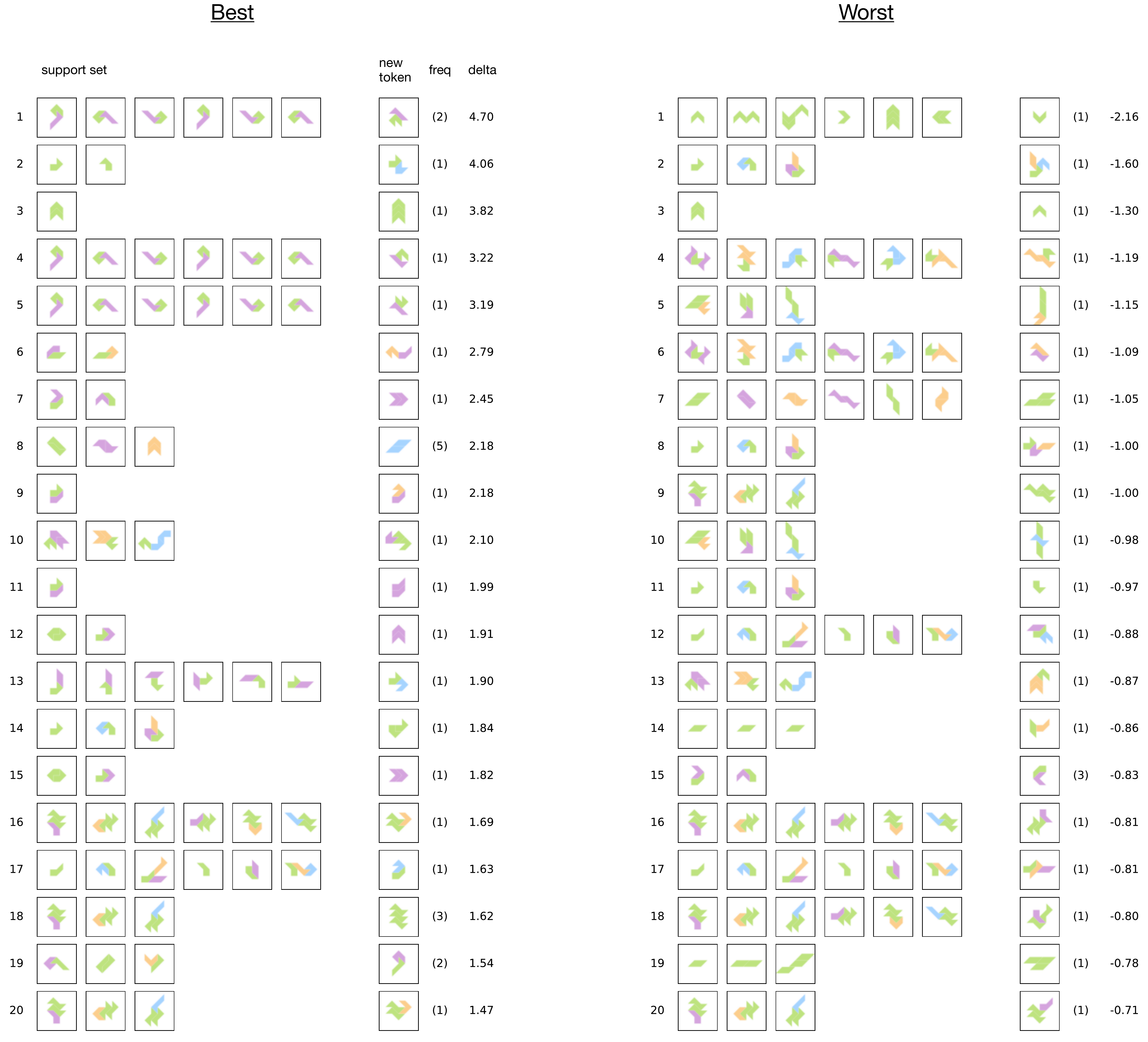}
    \caption{Best and worst 20 human examples, measured by $\ell$(GNS) - $\ell$(Bayes).}
    \label{fig:gns_vs_bayes}
\end{figure}

\begin{figure}[ht!]
    \centering
    \includegraphics[width=1.1\hsize]{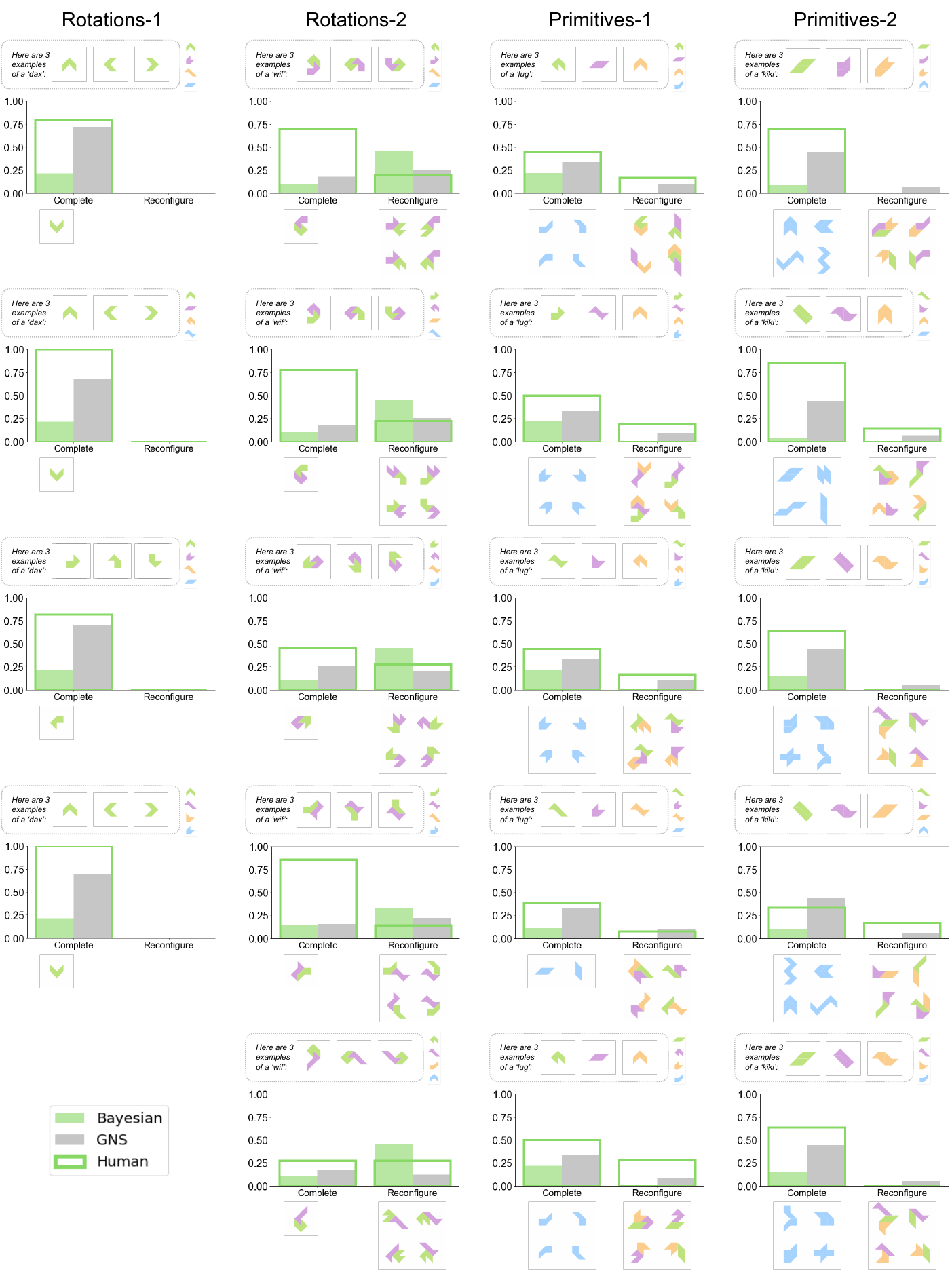}
    \caption{Inductive biases captured by GNS and Bayesian models (exhaustive version).}
    \label{fig:gns_inductive_biases}
\end{figure}

\end{document}